\documentclass[11pt]{article}
\usepackage{acl}
\usepackage{times}
\usepackage{latexsym}
\usepackage[T1]{fontenc}
\usepackage[utf8]{inputenc}
\usepackage{microtype}
\usepackage{inconsolata}
\usepackage{amsmath,amssymb,amsfonts}
\usepackage{booktabs}
\usepackage{array}
\usepackage{multirow}
\usepackage{graphicx}
\usepackage{colortbl} 
\usepackage[ruled]{algorithm}
\usepackage{algpseudocode}
\usepackage[most]{tcolorbox}
\renewcommand{\algorithmicrequire}{\textbf{Input:}}
\renewcommand{\algorithmicensure}{\textbf{Output:}}

\newcommand{\ours}{DSL-LLaDA} 
\newcommand{\best}[1]{\textbf{#1}}

\title{DSL-LLaDA: Scaling Continuous Denoising to 8B Masked Diffusion LMs}

\author{
  \textbf{Longxuan Yu\textsuperscript{1}\thanks{Equal contribution.}},
  \textbf{Yunshu Wu\textsuperscript{1}\footnotemark[1]},
  \textbf{Yu Fu\textsuperscript{1}},
  \textbf{Siheng Xiong\textsuperscript{2}},
\\
  \textbf{Rob Brekelmans},
  \textbf{Hui Liu\textsuperscript{3}},
  \textbf{Yue Dong\textsuperscript{1}},
  \textbf{Greg Ver Steeg\textsuperscript{1}\thanks{Corresponding author: \texttt{gregoryv@ucr.edu}}}
\\
\\
  \textsuperscript{1}University of California, Riverside,
  \textsuperscript{2}Georgia Institute of Technology,
  \textsuperscript{3}Microsoft
}

\begin{document}
\maketitle

\begin{abstract}
Discrete Masked diffusion language models generate text by iterative parallel decoding, but few-step decoding suffers from a tradeoff between length and quality: with a fixed step budget, standard methods can generate a short, high-quality output, or they can produce long but repetitive text. Continuous denoising can sidestep this tradeoff by evolving all positions jointly in embedding space, but building such a model from scratch at scale remains an open problem. We show that a pretrained masked DLM can instead be lightly adapted to support continuous embedding-space denoising. Starting from LLaDA-8B-Instruct, we continue-pretrain for only 1{,}000 steps with Discrete Stochastic Localization (DSL), replacing binary masking with continuous per-token Gaussian noise as a soft mask. The adapted model supports continuous inference that evolves all positions jointly in embedding space and defers hard token commitment to the final step. On zero-shot summarization at low step budgets ($\leq$16 forward passes), DSL-LLaDA-SDE achieves the best ROUGE-1 on all four benchmarks and largely avoids the premature-termination / repetition tradeoff of iterative unmasking. The same adaptation also yields selective noisy-state robustness: the model corrects corrupted tokens while preserving clean ones. Control experiments using standard masked diffusion training with the same compute demonstrate neither behavior.
\end{abstract}

\section{Introduction}
\label{sec:intro}

Masked diffusion language models (DLMs) have recently scaled to billions of parameters and begun to approach autoregressive quality on standard benchmarks~\cite{llada,dream,idlm}, yet their iterative unmasking inference degrades sharply at low step budgets, producing truncated outputs, repetition loops, and incoherent text~\cite{flowmap}. The root cause is that each position is decoded independently from its context-conditioned marginal distribution, neglecting inter-token correlations~\cite{flowmap}: this factorization introduces little error when each step changes only a few positions, but becomes severe at low step budgets, where each step must commit many positions at once.

\begin{figure*}[t!]
\centering
\includegraphics[width=\textwidth]{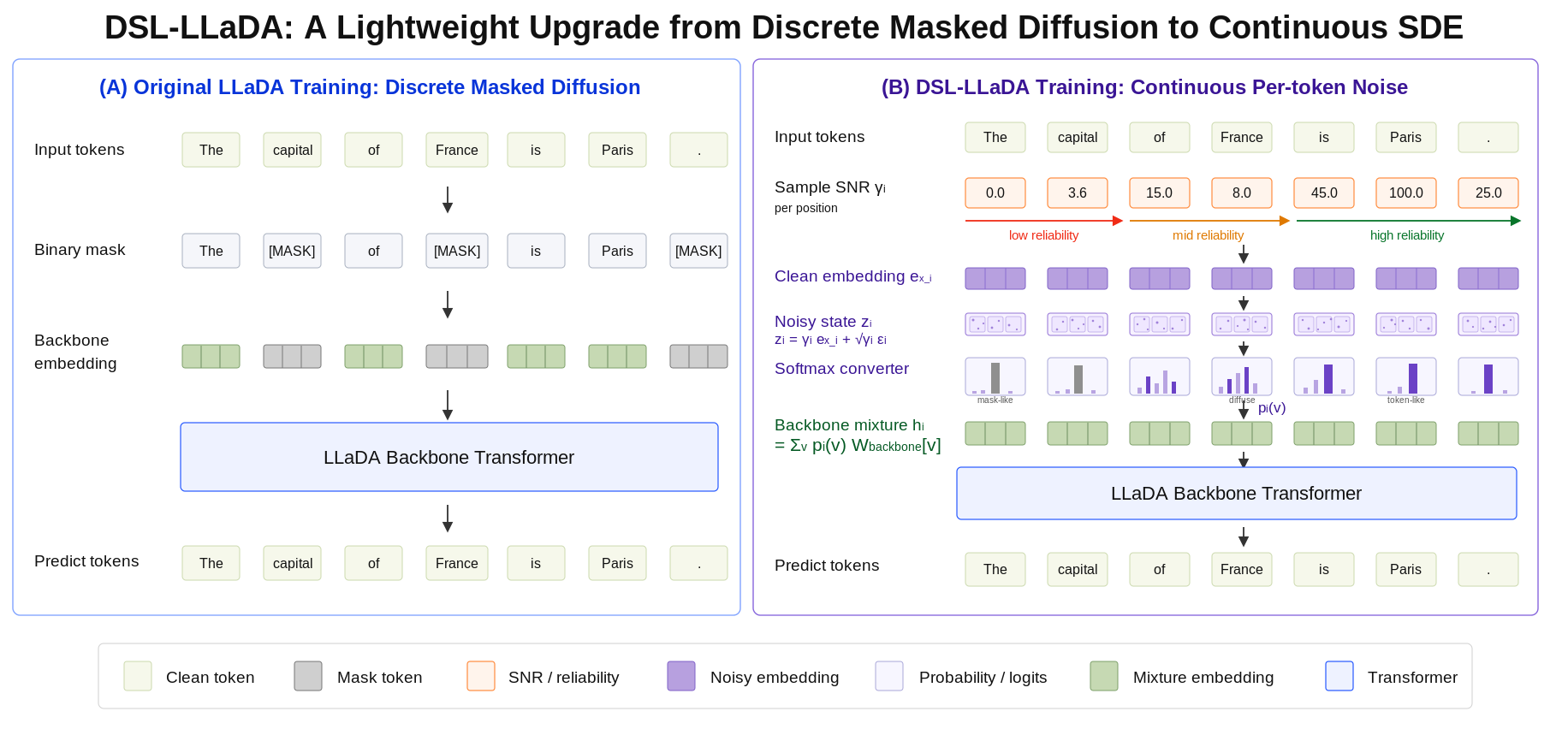}
\caption{DSL-LLaDA replaces binary masking with continuous per-token noise, enabling SDE-style continuous inference from a pretrained masked DLM. (A)~Original LLaDA uses discrete masked diffusion, where each token is either fully visible or replaced by a \texttt{[MASK]} token. (B)~DSL-LLaDA introduces continuous per-token noise at varying SNR. A softmax converter maps each noisy embedding back to the backbone's input space, producing three phases: unknown (mask-like at low SNR), unreliable (random tokens at mid SNR), and clear (gold token at high SNR), as shown by the per-position probability distributions.}
\label{fig:overview}
\end{figure*}


These failures motivate methods that improve few-step generation. Improved discrete samplers~\cite{remdm,prism,adlm} and consistency distillation~\cite{cd4lm,cdlm} reduce the step cost of discrete inference, but they still operate in the discrete token space. Continuous denoising offers a complementary route: it sidesteps the token-space factorization by evolving all positions jointly in a shared embedding space~\cite{flowmap,lipman2023flow}, and recent work confirms that continuous flows can match or exceed discrete diffusion in the few-step regime~\cite{flowmap,langflow}. However, building a continuous DLM from scratch at scale remains an open design problem, as embedding geometry, noise scheduling, and training objectives are still actively explored~\cite{gao2024embedding,langflow,flowmap}, and no continuous DLM has been trained beyond moderate scale.


Rather than training a continuous DLM from scratch, we show that an existing masked DLM can be lightly adapted to accept continuous inputs (Figure~\ref{fig:overview}). Starting from LLaDA-8B-Instruct, we continue-pretrain for only 1{,}000 steps with Discrete Stochastic Localization (DSL;~\citealt{dsl2}), replacing the binary clean-or-mask input with a soft mixture over token embeddings that encodes graded per-position uncertainty. The resulting model, \ours{}, supports a stochastic differential equation (SDE) based continuous denoising sampler with delayed hard token commitment.

Empirically, the adapted model shows three behaviors absent in iterative unmasking. On open-ended generation, the SDE sampler's delayed token commitment keeps repetition below 9\% at all step budgets, while discrete baselines either truncate early or exceed 60\% repetition (Section~\ref{sec:sde}). On zero-shot summarization, continuous inference achieves the best ROUGE-1 on all four benchmarks at $\leq$16 forward passes without length-control heuristics (Section~\ref{sec:sde}). On error correction, the model corrects most corrupted tokens while leaving ${>}$98\% of clean tokens untouched, whereas XDLM-style random-token training preserves fewer than half (Section~\ref{sec:correction}). A same-compute binary-masking control experiment acquires none of these behaviors, confirming that continuous-noise exposure, not additional training alone, drives the improvement.

Our contributions are:
\begin{enumerate}
\item We show that a pretrained 8B masked DLM can be converted to support continuous-space denoising through lightweight DSL adaptation.
\item We demonstrate that continuous SDE inference avoids the low-NFE failure modes of iterative unmasking on generation and summarization tasks.
\item We identify a selective noisy-state robustness induced by continuous-noise training, absent from same-compute discrete baselines.
\end{enumerate}

\section{Background}
\label{sec:background}

\subsection{Masked Diffusion Language Models}

\paragraph{Notation.} Let $x_{1:L}$ denote a token sequence over a vocabulary of size $V$. The backbone maps each token to an embedding 
via a learned matrix $W_{\text{backbone}} \in \mathbb{R}^{(V+1) \times d_{\text{model}}}$, where the extra row corresponds to a dedicated \texttt{[MASK]} token.

\paragraph{Training.} A masked diffusion language model defines a forward process that independently replaces each token with \texttt{[MASK]}. At diffusion step $t$, position $i$ is either fully visible ($x_i^t = x_i$) or fully masked ($x_i^t = \texttt{[MASK]}$). No intermediate state exists. The model is trained to predict the original token at every position using cross-entropy loss.

\paragraph{Inference.} Generation starts from a fully masked sequence and proceeds by iterative \emph{unmasking}: at each of $T$ steps the model predicts token probabilities at all masked positions in parallel, samples tokens from these distributions, and commits (unmasks) a subset of positions. Uncommitted positions are re-masked for the next step. This generative process induces two sources of error relevant to this work: (1)~each position is decoded independently given the current state, factorizing the reverse transition across positions and neglecting inter-token correlations~\cite{flowmap}, and (2)~sampling tokens from an imperfect model creates a mismatch with the training data, which cannot be corrected after unmasking.

\subsection{Continuous Denoising for Discrete Tokens}

An alternative to iterative unmasking is to maintain a continuous state for each token position throughout generation~\cite{plaid,dsl2}. Every position holds a soft, noisy representation rather than a committed token, and a denoiser iteratively recovers clean-token estimates until hard tokens are read out at the final step. Since the denoiser conditions on the full noisy sequence, the update is coupled across positions, avoiding the independent-position error described above~\cite{langflow}. However, continuous denoising requires a backbone trained to process soft inputs, a capability that standard masked diffusion models lack.

When the continuous state lives in token embedding space, Stochastic Localization~\cite{dsl2} offers a natural formulation: as a per-token signal-to-noise ratio (SNR) increases from zero, the model's belief about which token occupies each position sharpens from uniform (completely uncertain) to concentrated on the correct token. Constraining the token embeddings to the unit sphere makes the optimal denoiser depend only on the noisy state and not on explicit parameters such as the SNR~\cite{dsl2}. This SNR-invariance has two practical consequences: the model does not need an explicit timestep input, which is what makes it possible to adapt a pretrained masked DLM without any architectural changes; and each token position may follow its own SNR schedule, which we exploit during training (\S\ref{sec:dsl_adaptation}) and which gives the associated generative SDE a time-free drift (\S\ref{sec:normalized_sampling}). In Section~\ref{sec:method} we show how to adapt a pretrained masked DLM to serve as the denoiser in this framework.

\section{Method}
\label{sec:method}

We adapt LLaDA-8B-Instruct to process continuous DSL states and then use the adapted model as a denoiser for continuous response generation.
The training and sampling parts of the method are conceptually separate: during training, we expose the backbone to noisy token states and train with the standard token cross-entropy; during sampling, we maintain a continuous response state, query the adapted backbone for a clean-token estimate, and update the state without committing to hard tokens until the final step.

Throughout this section, $x_i$ denotes the clean token at position $i$ and $\gamma$ denotes SNR. The method involves two separate embedding matrices: $W_{\text{token}} \in \mathbb{R}^{(V+1) \times d}$, a learned noise embedding table whose rows $e_v \in \mathbb{R}^d$ ($d{=}100$) are constrained to the unit sphere and used to construct DSL noise states; and $W_{\text{backbone}} \in \mathbb{R}^{(V+1) \times d_{\text{model}}}$, initialized from the pretrained LLaDA input embedding matrix (including the mask row) and trainable during DSL adaptation. The converter bridges these two spaces.

\subsection{DSL Adaptation}
\label{sec:dsl_adaptation}

DSL replaces binary clean-or-mask inputs with continuous noisy states that encode how much information each position carries about its token. For a clean token $x_i$, we sample an SNR value $\gamma_i \geq 0$ and construct
\[
z_i = \gamma_i \cdot e_{x_i} + \sqrt{\gamma_i} \cdot \epsilon_i, \quad \epsilon_i \sim \mathcal{N}(0, I_d).
\]
At $\gamma_i{=}0$ this gives $z_i{=}0$, equivalent to a fully masked position. As $\gamma_i$ increases, the state carries progressively more information about the clean token.

The LLaDA backbone expects token-like input embeddings, so we map each DSL state $z_i$ through a softmax converter:
\[
h_i = \operatorname{softmax}\bigl(\beta \cdot z_i^\top W_{\text{token}}^\top + b^\top\bigr)\, W_{\text{backbone}}
\]

where $\beta$ is a learnable temperature parameter that controls the sharpness of the softmax distribution. $W_{\text{token}} \in \mathbb{R}^{(V+1) \times d}$ reuses the noise embedding table for its first $V$ rows and appends a learned mask embedding for the $(V{+}1)$-th slot (initialized from the pretrained mask embedding and trainable). The zero-SNR state maps to a mask-like input, while high-SNR states become increasingly token-like, converging to the corresponding row of $W_{\text{backbone}}$, which is initialized from the pretrained embedding (Figure~\ref{fig:overview}B; detailed converter behavior in Appendix~\ref{app:converter}).

Given the converted inputs $h_{1:L}$, the backbone predicts token logits at all positions. We train with the standard token cross-entropy:
\[
\mathcal{L} = -\sum_{i=1}^{L} \log p_{\theta,i}(x_i \mid h_{1:L}).
\]
The target is always the original clean token. Thus DSL adaptation changes the input states seen by the backbone, but not the prediction target.

SNR values $\gamma_i$ are drawn from a mixed distribution that combines continuous noise levels with mask-like endpoint states, so the model sees a range of inputs from fully uncertain to fully clean during training. We continue-pretrain LLaDA-8B-Instruct for 1{,}000 steps on FineWeb-Edu. The SNR distribution, logit scale $\beta$, and all other hyperparameters are detailed in Appendix~\ref{app:details} and \ref{app:design_space}.

\subsection{Clean-Token Embedding Estimate}
\label{sec:posterior_mean_denoiser}

After DSL adaptation, the backbone predicts a probability distribution over tokens at each position. For sampling, we convert this distribution into a single embedding vector that represents the model's best guess of the clean token. Let $S_i$ be the top-$M$ most likely tokens under $p_{\theta,i}(\cdot)$. We renormalize over this set and compute a weighted average of their embeddings:
\[
\hat{e}_i = \sum_{v \in S_i} \tilde{p}_{\theta,i}(v) \cdot e_v, \quad \tilde{p}_{\theta,i}(v) = \frac{p_{\theta,i}(v)}{\sum_{u \in S_i} p_{\theta,i}(u)}.
\]
This is the only model output used by the continuous sampler. We use $M{=}512$ in all experiments.

\subsection{Continuous Sampling}
\label{sec:normalized_sampling}

The DSL localization channel admits unconditional generative dynamics whose drift is the posterior-mean denoiser~\cite{dsl2}:
\[
\mathrm{d}z_i = \hat{x}_i(z)\,\mathrm{d}t + \mathrm{d}W_i,
\]
Because token embeddings lie on the unit sphere, the denoiser $\hat{x}_i(z)$ depends only on $z$ and not on $\gamma_i$ (Section~\ref{sec:background}), so the drift has no explicit time dependence and each token may follow its own SNR path.

We sample by advancing the normalized state $y_i = z_i/\gamma_i$ with Heun steps over an increasing SNR schedule $\gamma_0 < \cdots < \gamma_{N/2}$. At each step, the current state is converted to a DSL input $z_i = \gamma \cdot y_i$, passed through the converter and backbone, and the clean-token embedding estimate $\hat{e}_i$ from Section~\ref{sec:posterior_mean_denoiser} is used to update $y$:
\[
y_i \;\leftarrow\; y_i + \Delta\gamma \cdot \frac{\hat{e}_i - y_i}{\gamma}.
\]
Each Heun step queries the backbone twice (predict then correct), and no hard token decisions are made until the final step (Algorithm~\ref{alg:inference}). The full derivation from the SDE to this update rule, along with the SNR schedule and other sampler details, are in Appendix~\ref{app:details} and \ref{app:sde_sensitivity}.

\subsection{Conditional Generation and Decoding}
\label{sec:conditional_generation}

For instruction following and summarization, we concatenate a context $c_{1:m}$ (prompt or source document) with a response region of length $L$. Context positions are clamped to their clean backbone embeddings throughout the trajectory; the continuous update runs only on response positions. Because the backbone attends bidirectionally to the full sequence, each denoiser call is conditioned on both the clean context and the current noisy response state. At the final SNR $\gamma_{N/2}$, we decode by $\hat{x}_i = \arg\max_v p_{\theta,i}(v)$ and truncate at the first EOS token.

\section{Experiments}
\label{sec:experiments}

DSL adaptation expands the input states seen by LLaDA from binary clean/mask tokens to continuous uncertain token mixtures. We evaluate this expanded state support in two complementary ways: (1)~noisy-state probes test whether the adapted backbone can distinguish unreliable from reliable context (\S\ref{sec:robustness}), and (2)~generation experiments test whether the same adapted backbone can use an SDE-style continuous sampler to avoid the low-NFE failure modes of iterative unmasking (\S\ref{sec:sde}).

\paragraph{Setup.} All experiments share LLaDA-8B-Instruct~\cite{llada} as the backbone. DSL-LLaDA is continue-pretrained for 1{,}000 steps on FineWeb-Edu with $\beta{=}1$ and random unit-norm noise embeddings. We compare against five baselines: \textbf{LLaDA} (default iterative unmasking); \textbf{LLaDA$^\star$} (our strongest length-controlled discrete setting, with EOS suppression and block remasking; Appendix~\ref{app:discrete_tuning}); \textbf{MDM-CPT} and \textbf{XDLM}~\cite{xdlm} (same backbone, data, steps, and compute as DSL-LLaDA, but with standard binary masking and random-token corruption respectively, isolating the effect of continuous-noise exposure); and \textbf{LLaDA ReMDM}~\cite{remdm} (confidence-based remasking applied to the original LLaDA weights at inference only, without additional training, so its remaining gap to DSL-LLaDA indicates that discrete rescheduling alone is insufficient). Context-robustness and error-correction evaluations use 100 texts from WikiText-103; generation and summarization setups are in \S\ref{sec:sde}. Full hyperparameters appear in Appendix~\ref{app:details}.

\subsection{Context Robustness and Selective Correction}
\label{sec:robustness}

We probe whether DSL adaptation gives the backbone selective noisy-state capability, then show that it also enables correction of corrupted tokens.

\begin{figure}[t!]
\centering
\includegraphics[width=\columnwidth]{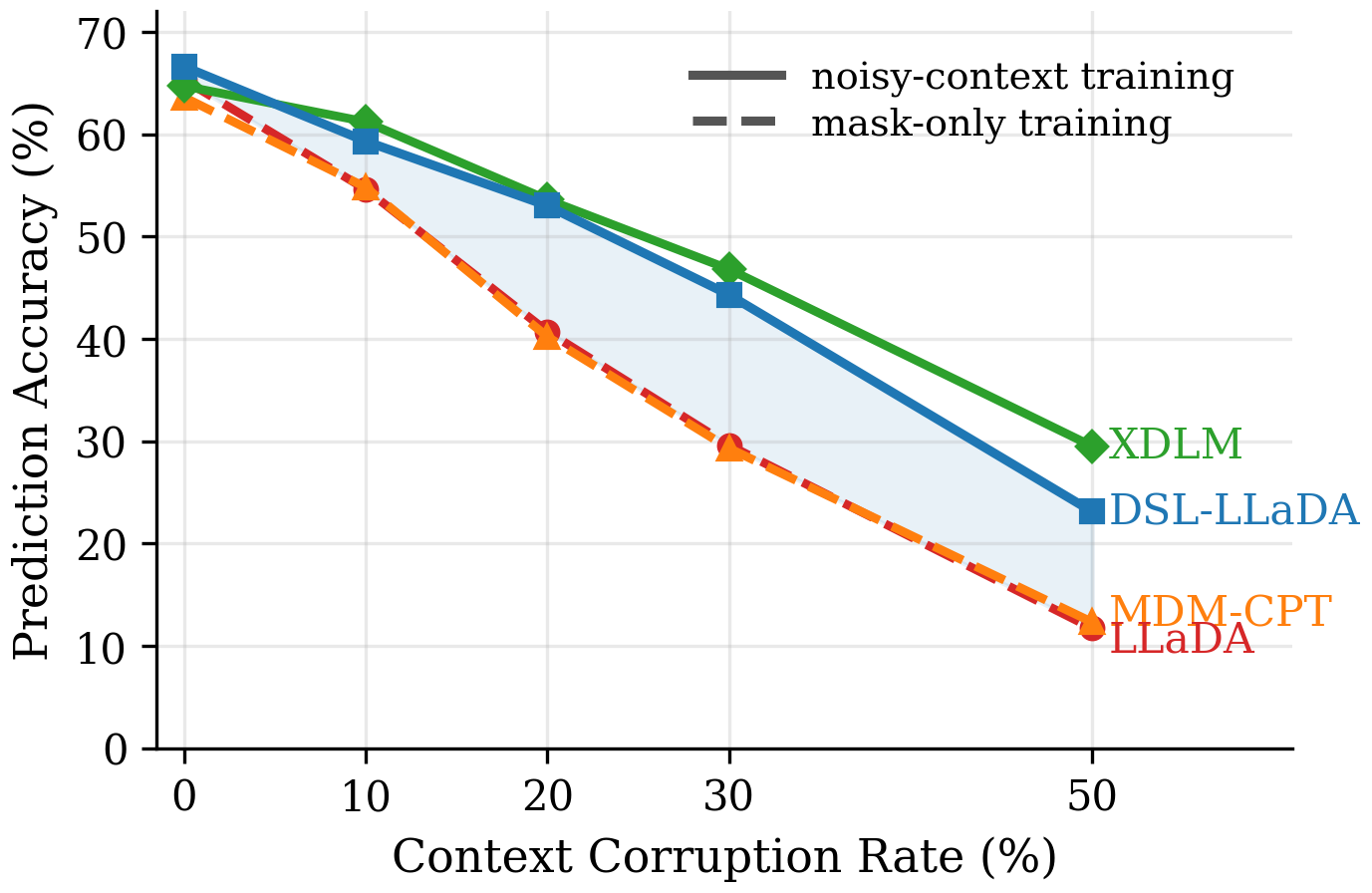}
\caption{Context robustness (mask=30\%, 100 texts). Models trained with noisy context (solid) degrade at roughly half the rate of mask-only models (dashed). Training with corruption determines robustness; the type of corruption (continuous Gaussian vs.\ discrete random-token) matters less.}
\label{fig:context_robust}
\end{figure}

\paragraph{Robustness to corrupted context.}
We mask 30\% of tokens, corrupt $c$\% of the remaining visible context with random tokens, and measure masked-position accuracy (Figure~\ref{fig:context_robust}).

As Figure~\ref{fig:context_robust} shows, models trained with noisy context (XDLM, DSL-LLaDA) degrade at roughly half the rate of mask-only models (LLaDA, MDM-CPT), which overlap almost perfectly despite different training durations. The gap widens with corruption: at 50\% corruption, noisy-context models retain 23--30\% accuracy while mask-only models fall to ${\sim}$12\%. Notably, MDM-CPT tracks LLaDA rather than the noisy-context models, confirming that robustness stems from exposure to corrupted context during training, not from additional pretraining compute.

\paragraph{Selective correction of noisy tokens.}
\label{sec:correction}
This robustness has a practical consequence: the model can identify and correct corrupted tokens rather than reproducing them unchanged. Table~\ref{tab:corruption} measures two complementary metrics: \emph{Fix}, the percentage of corrupted positions correctly recovered, and \emph{Clean}, the percentage of uncorrupted positions that the model leaves unchanged.

\begin{table}[t!]
\centering
\caption{Random-token error correction (single forward pass, 100 texts). Fix = corrupted positions correctly predicted; Clean = uncorrupted positions preserved. XDLM achieves the highest fix rate yet damages ${\sim}$1 in 5 clean tokens at 30\% corruption. DSL-LLaDA ($\beta{=}1$) preserves ${>}$98\% clean tokens at every level.}
\label{tab:corruption}
\resizebox{\columnwidth}{!}{%
\begin{tabular}{l cc cc cc}
\toprule
& \multicolumn{2}{c}{\textbf{Random @10\%}} & \multicolumn{2}{c}{\textbf{Random @30\%}} & \multicolumn{2}{c}{\textbf{Random @50\%}} \\
\cmidrule(lr){2-3}\cmidrule(lr){4-5}\cmidrule(lr){6-7}
\textbf{Model} & Fix & Clean & Fix & Clean & Fix & Clean \\
\midrule
LLaDA              & 13.1 & 90.9 & 5.9 & 89.3 & 1.1 & 87.4 \\
MDM-CPT            & 17.2 & 92.0 & 10.4 & 90.6 & 3.5 & 87.3 \\
XDLM               & \best{70.8} & 82.0 & \best{55.7} & 71.0 & \best{33.3} & 45.7 \\
\ours{}\,($\beta{=}1$, ours) & 64.3 & \best{99.0} & 48.5 & \best{98.7} & 21.5 & \best{98.1} \\
\bottomrule
\end{tabular}}
\end{table}

\ours{} selectively corrects corrupted tokens while preserving 98\%+ clean tokens. MDM-CPT acquires only marginal error correction, with fix rates barely exceeding LLaDA's. XDLM achieves the highest fix rate yet damages clean tokens severely, preserving only 45\% of clean positions at 50\% corruption. DSL-LLaDA at $\beta{=}1$ closes most of the fix-rate gap while maintaining clean-token preservation above 98\% at every corruption level. A direct measure of selectivity is the number of corrupted tokens corrected per clean-token error, Fix$/(100-$Clean$)$: at 50\% corruption \ours{} obtains 11.3, compared with 0.61 for XDLM. Selective correction does not compromise the original mask-filling objective: at 30\% masking the $\beta{=}1$ model reaches 67.0\% accuracy vs.\ LLaDA's 65.1\%, and ECE \emph{improves} (0.026 vs.\ 0.039; Appendix~\ref{app:mask_calibration}).

On random-replacement corruptions, DSL variants match XDLM's fix rate while preserving $>$98\% of clean tokens. On semantic or factual substitutions, the primary $\beta{=}1$ model preserves context rather than rewriting. We therefore frame DSL-LLaDA as a selective noisy-context model, not a general factual corrector (Appendix~\ref{app:design_space}).

\begin{figure*}[t!]
\centering
\includegraphics[width=\textwidth]{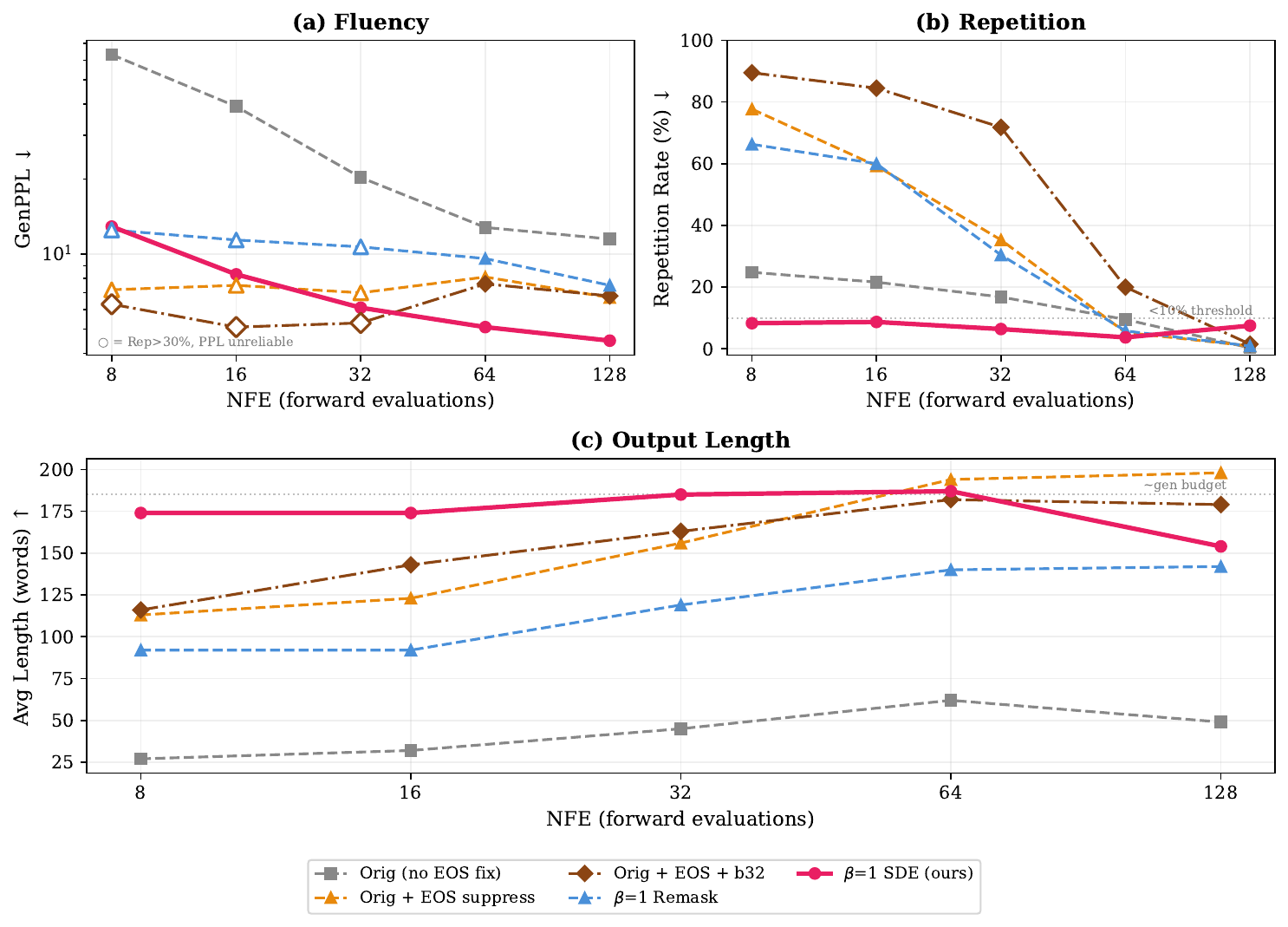}
\caption{DSL-LLaDA-SDE avoids the low-NFE failure modes of iterative unmasking, maintaining sub-10\% repetition and full-length outputs at all step budgets (200 prompts, gen=256).
\textbf{(a)}~Fluency (perplexity; hollow markers indicate unreliable estimates due to high repetition).
\textbf{(b)}~Repetition: DSL-LLaDA-SDE stays below 10\% at all NFE, while length-controlled discrete variants exceed 60\% at NFE${\leq}$16 (vanilla LLaDA avoids repetition only by terminating early).
\textbf{(c)}~Output length: vanilla LLaDA truncates to $<$60 words due to premature termination, whereas DSL-LLaDA-SDE consistently produces $>$150 words.}
\label{fig:nfe_curve}
\end{figure*}

\paragraph{Summary.}
DSL adaptation gives the backbone a selective noisy-state capability that does not arise from binary continued pretraining (MDM-CPT tracks LLaDA despite matched compute) and is qualitatively different from XDLM-style random-token corruption (DSL-LLaDA preserves clean tokens far better than XDLM at comparable fix rates).

\subsection{Continuous Generation via SDE}
\label{sec:sde}

\paragraph{Setup.} Open-ended generation uses 200 instruction-style prompts (gen=256); long-form generation extends to 512 and 1{,}024 tokens. Zero-shot summarization uses 1{,}000 examples (seed 42) from XSum~\cite{xsum} (21w, gen$=$128), CNN/DailyMail~\cite{cnndm} (55w, gen$=$256), arXiv~\cite{pubmed_summ} (163w, gen$=$256), and PubMed~\cite{pubmed_summ} (205w, gen$=$256). Metrics include perplexity (GPT-2 Large), repetition rate (Rep: fraction of adjacent word pairs that are identical; Appendix~\ref{app:details}), ROUGE F1, BERTScore, and LLM-as-judge (GPT-5.4). Full metric definitions and dataset sourcing appear in Appendix~\ref{app:details}.

\begin{table*}[t]
\centering
\caption{Zero-shot summarization ROUGE-1 F1 across NFE budgets (1{,}000 samples per benchmark, seed=42). SDE leads at every NFE on XSum and CNN/DM, and at NFE$\leq$16 on all four benchmarks; LLaDA reaches comparable scores on long-reference tasks at NFE$\geq$32 once sufficient refinement steps suppress its premature-termination artifacts. LLaDA$^\star$ underperforms vanilla LLaDA on long references and produces 7--25\% degenerate endings on short references (Appendix~\ref{app:sum_failures}, \ref{app:block_baseline}). LLaDA ReMDM improves over vanilla LLaDA at low NFE but still suffers from premature termination at high NFE. A dev sweep confirms that tuning does not remove this tradeoff (Appendix~\ref{app:discrete_tuning}). AR references and full metric tables: Appendix~\ref{app:ar_baseline}.}
\label{tab:summarization_hero}
\footnotesize
\setlength{\tabcolsep}{6pt}
\renewcommand{\arraystretch}{1.05}
\begin{tabular}{ll cccc}
\toprule
Dataset (ref) & Method & NFE=8 & 16 & 32 & 64 \\
\midrule
\multirow{4}{*}{XSum (21w)}
& \textbf{DSL-LLaDA-SDE} (ours) & \best{28.4} & \best{30.4} & \best{32.0} & \best{32.4} \\
& LLaDA                         & 25.2        & 29.0        & 28.8        & 24.8        \\
& LLaDA ReMDM                    & 24.9        & 27.8        & 25.6        & 14.9        \\
& LLaDA$^\star$                  &  9.9        & 14.1        & 20.2        & 24.1        \\
\midrule
\multirow{4}{*}{CNN/DM (55w)}
& \textbf{DSL-LLaDA-SDE} (ours) & \best{28.1} & \best{33.0} & \best{35.1} & \best{35.8} \\
& LLaDA                         & 23.1        & 28.2        & 25.4        & 18.2        \\
& LLaDA ReMDM                    & 23.3        & 27.1        & 20.9        &  9.2        \\
& LLaDA$^\star$                  &  8.6        &  8.3        & 18.7        & 28.4        \\
\midrule
\multirow{4}{*}{PubMed (205w)}
& \textbf{DSL-LLaDA-SDE} (ours) & \best{29.4} & \best{32.2} & \best{36.9} & 39.5        \\
& LLaDA                         & 11.9        & 20.2        & 36.5        & \best{42.2} \\
& LLaDA ReMDM                    & 11.5        & 19.7        & 36.3        & \best{42.2} \\
& LLaDA$^\star$                  &  6.1        &  7.5        & 20.9        & 37.8        \\
\midrule
\multirow{4}{*}{arXiv (163w)}
& \textbf{DSL-LLaDA-SDE} (ours) & \best{27.3} & \best{28.6} & 32.9        & 34.9        \\
& LLaDA                         & 10.6        & 16.2        & 33.8        & 40.8        \\
& LLaDA ReMDM                    & 10.3        & 15.9        & \best{34.3} & \best{41.0} \\
& LLaDA$^\star$                  &  7.8        &  6.6        & 15.9        & 34.7        \\
\bottomrule
\end{tabular}
\end{table*}

\paragraph{NFE-efficiency and failure modes.}
We track perplexity, repetition, and output length jointly because they trade off against each other: a model can achieve low perplexity by repeating fluent phrases, low repetition by terminating early, or long output by looping. Only a method that scores well on all three avoids degenerate generation.
Iterative unmasking exhibits three failure modes that Figure~\ref{fig:nfe_curve} traces across NFE budgets: premature termination (truncated outputs), token repetition loops, and degenerate endings. At NFE$\leq$32, default iterative unmasking avoids repetition only by terminating early, while length-controlled discrete variants exceed 60\% repetition at NFE$\leq$16. DSL-LLaDA-SDE stays below 9\% without either failure mode. At NFE$=$64, DSL-LLaDA-SDE reaches perplexity 5.1 at 3.7\% repetition. At NFE$=$128, discrete methods close the repetition gap and DSL-LLaDA-SDE's repetition slightly exceeds the best discrete variant, though DSL-LLaDA-SDE retains a perplexity advantage (4.5 vs.\ 6.7 for LLaDA+EOS).
The DSL-trained model performs \emph{worst} under iterative unmasking at low NFE (60--66\% repetition at NFE$\leq$16), confirming that DSL optimizes the backbone for continuous-space inputs. DSL-LLaDA-SDE is the inference mode the DSL-trained backbone requires (Appendix~\ref{app:gen_quality}).

\paragraph{Long-form generation.}
The repetition problem worsens with generation length: at 512 tokens, length-controlled LLaDA variants exceed 68\% repetition and DSL-LLaDA under iterative unmasking reaches 43\%, while DSL-LLaDA-SDE stays at 11\%. At 1{,}024 tokens the pattern persists with DSL-LLaDA-SDE at 28\% (Appendix~\ref{app:longform}).

\paragraph{Zero-shot summarization.}
ROUGE-1 F1 across NFE budgets appears in Table~\ref{tab:summarization_hero}. BERTScore appears in Appendix~\ref{app:sum_bertscore}; failure-mode rates and length statistics are in Appendix~\ref{app:sum_failures}; R-2/R-L against AR baselines are in Appendix~\ref{app:ar_baseline}.

The strongest regime for DSL-LLaDA-SDE is low-NFE anchored generation. Four patterns emerge: (1)~SDE leads R-1 at NFE$\leq$16 on all four benchmarks and at every NFE on XSum and CNN/DM, with scores improving monotonically, whereas LLaDA peaks at NFE$=$16 on CNN/DM and decays to 18.2 at NFE$=$64 as premature termination worsens. (2)~LLaDA ReMDM tracks vanilla LLaDA closely but does not close the gap to SDE: on short-reference tasks (XSum, CNN/DM), RMDM peaks at NFE$=$16 and then collapses even more severely than vanilla LLaDA (14.9 and 9.2 R-1 at NFE$=$64), confirming that confidence-based remasking does not resolve the premature-termination problem. On long-reference tasks, RMDM matches vanilla LLaDA at all NFE budgets. (3)~LLaDA$^\star$ becomes competitive only at NFE$\geq$32 and still lags vanilla LLaDA on long references at NFE$=$64 (Appendix~\ref{app:discrete_tuning} tunes iterative unmasking on development subsets and finds the same tradeoff). (4)~The trend reverses on long references: at NFE$=$64, LLaDA leads SDE on PubMed ($-$2.7 R-1) and arXiv ($-$5.9) once outputs reach reference length. SDE matches reference lengths without any length control and emits $\leq$4\% degenerate endings, while LLaDA's short-reference outputs collapse to 12--14 words (Appendix~\ref{app:sum_failures}). Compared to same-scale AR models (Appendix~\ref{app:ar_baseline}), SDE exceeds both Qwen2.5-7B and Llama-3.1-8B by 4--6 R-1 on XSum; on long references the AR models lead by 2--10 R-1.

\paragraph{LLM-as-judge evaluation.}
We complement ROUGE with pairwise preference judgments (GPT-5.4, 100 samples/dataset, positions randomized; Table~\ref{tab:sum_quality} in Appendix~\ref{app:judge_axes}). At NFE=16, the judge prefers SDE over vanilla LLaDA in aggregate (mean 57.9\% vs.\ 26.6\%; SDE wins three of four benchmarks, with CNN/DM nearly tied at 45.7\% vs.\ 46.9\%) and decisively over LLaDA$^\star$ (mean 83.6\% vs.\ 1.5\%). At NFE=64, SDE retains its lead over LLaDA$^\star$ (mean 59.8\% vs.\ 29.4\%). Vanilla LLaDA wins only on long references at high NFE where premature termination no longer occurs.

Summarization aligns with SDE's strengths: the source document anchors the denoising trajectory to source semantics, reducing the attractor collapse (all positions converging on high-frequency tokens) observed in open-ended creative generation (24\% degenerate on creative prompts vs.\ $\leq$4\% on summarization; Appendix~\ref{app:examples}). SDE produces well-structured outputs without length control or termination heuristics. Wall-clock measurements appear in Appendix~\ref{app:ar_baseline}.

\section{Related Work}

\textbf{Diffusion language models.}
Discrete DLMs~\cite{llada,dream,mdlm,md4,sedd} train on clean context and use iterative unmasking at inference. Continuous approaches~\cite{diffusionlm,cdcd,plaid,ssdlm,langflow,flowmap} operate in embedding or simplex space but have not been scaled beyond moderate size. Soft-MDLM~\cite{softmdlm} replaces hard masking with soft mixtures but does not include SDE generation. BD3-LMs~\cite{bd3lm} and Mercury~\cite{mercury} combine autoregressive blocks with within-block discrete diffusion. Our work does not propose a new diffusion framework; instead we ask whether an existing 8B masked DLM can be lightly converted to support continuous inference via DSL~\cite{dsl2}.

\textbf{Noisy-context training.}
Several methods expose masked DLMs to noisy rather than clean context during training. Scheduled sampling~\cite{scheduled_sampling} addresses the train-test gap for autoregressive models; XDLM~\cite{xdlm} adapts this idea to MDLMs by replacing tokens with random vocabulary words; Soft-MDLM~\cite{softmdlm} uses soft token mixtures instead of hard masks; EvoToken-DLM~\cite{evotoken} replaces hard masks with evolving soft token distributions for progressive refinement. DSL~\cite{dsl2} takes a different approach by adding continuous Gaussian noise in embedding space. Our work scales DSL to 8B and evaluates whether the converted backbone supports effective low-NFE continuous inference.

\textbf{Inference-time improvements.}
ReMDM~\cite{remdm}, PRISM~\cite{prism}, ADLM~\cite{adlm}, CD4LM~\cite{cd4lm}, and CDLM~\cite{cdlm} speed up discrete inference without changing the factorized transition structure. We include LLaDA ReMDM as an inference-only baseline in our summarization experiments (Table~\ref{tab:summarization_hero}); it tracks vanilla LLaDA closely and does not close the gap to SDE at low NFE. DSL instead changes the training objective to enable continuous generation; the two approaches are complementary.

\section{Discussion}

\paragraph{Evaluating DLMs beyond single metrics.}
Standard generation benchmarks report a single quality score such as perplexity or MAUVE, each inflatable by a different failure mode: early termination lowers perplexity while looping inflates output length. Our three-panel evaluation (Section~\ref{sec:experiments}) detects these pathologies jointly; no existing DLM benchmark explicitly checks for all three, which may explain why the length--quality tradeoff at low NFE has received limited attention.

\paragraph{Per-token time allocation and effective parallelism.}
Existing continuous DLMs assign the same SNR schedule to every position, implicitly treating all tokens as equally difficult. In practice, function words are low-entropy decisions while content-bearing tokens need more refinement. Discrete MDLMs can partially address this through confidence-based unmasking that resolves easier tokens first, though the per-step capacity remains shared. DSL's SNR-invariant denoiser (Section~\ref{sec:background}) removes the timestep dependence, letting each position follow its own SNR path in principle. A related question is how many tokens can be effectively denoised in parallel. SDE performs best on XSum, where summaries are short and tightly source-constrained, and the advantage narrows on longer targets, suggesting that effective parallelism depends on both model capability and task difficulty. We leave adaptive per-token scheduling and adaptive parallel block sizing to future work.

\section{Conclusion}

We show that a pretrained 8B masked DLM can be converted to support continuous embedding-space inference through 1{,}000 steps of DSL adaptation. The converted model achieves the best ROUGE-1 on all four summarization benchmarks at NFE$\leq$16 while avoiding the length--quality tradeoff of iterative unmasking, and acquires selective noisy-state robustness absent from same-compute discrete controls. These results suggest that continuous-noise adaptation is a general-purpose conversion procedure for masked DLMs.

\newpage
\section*{Limitations}

All results are on LLaDA-8B; other MDLMs and model scales are untested. SDE generation is sensitive to converter sharpness ($\beta{=}1$ works best; higher $\beta$ degrades). We scope evaluation to anchored generation; open-ended creative writing shows higher degeneration (24\% vs.\ $\leq$4\% on summarization). On long-reference tasks at high NFE, vanilla LLaDA matches or exceeds SDE. We do not claim parity with same-scale AR models (Appendix~\ref{app:ar_baseline}). Human evaluation remains future work.

\bibliography{references}

\newpage
\appendix
\section{Algorithms}
\label{app:algorithms}

Algorithm~\ref{alg:training} shows one DSL training step. Algorithm~\ref{alg:inference} shows the stochastic Heun-solver inference loop used in the main experiments.

\paragraph{Sampler derivation.}

\paragraph{Forward density.} The forward process defines a per-position Gaussian conditional:
\begin{align*}
 z_i \mid e_{x_i} & \;\sim\; \mathcal{N}\!\bigl(\gamma_i \cdot e_{x_i},\;\gamma_i \cdot I_d\bigr),
\\
 p_\gamma(z \mid e_v) &= (2\pi\gamma)^{-d/2}\exp\!\Bigl(-\frac{\|z - \gamma e_v\|^2}{2\gamma}\Bigr).
\end{align*}
The marginal over the token distribution $p(v)$ is $p_\gamma(z) = \sum_v p(v)\,p_\gamma(z \mid e_v)$.

\paragraph{Denoiser.} The optimal denoiser is the posterior mean:
\[
\hat{e} \;=\; \mathbb{E}[e_x \mid z] \;=\; \sum_v p(v \mid z)\,e_v,
\]
where the posterior weights are $p(v \mid z) \propto p(v)\,\exp\!\bigl(-\|z - \gamma e_v\|^2/(2\gamma)\bigr)$. Writing $y = z/\gamma$, the denoiser residual $\hat{e} - y$ points from the current normalized state toward the posterior mean.

\paragraph{Normalized state.} We reparameterize to $y = z/\gamma$. At SNR $\gamma$, $y = e_x + \epsilon/\sqrt{\gamma}$: as $\gamma$ increases from $\gamma_0$ to $\gamma_{\max}$, the noise contribution shrinks and $y$ converges to $e_x$. Generation proceeds by advancing $\gamma$ from low to high.

\paragraph{Update rule.} Our sampler uses the $\gamma$-time vector field
\[
v_y(y,\gamma) = \frac{\hat{e} - y}{\gamma}.
\]
At each step, the denoiser estimates the clean embedding $\hat{e}$ and the vector field nudges the current state $y$ toward it, with $1/\gamma$ scaling the step size relative to the current noise level. The Heun solver discretizes this vector field with a predictor-corrector scheme (Algorithm~\ref{alg:inference}). Our empirical claims do not require interpreting this update as an exact reverse-time SDE; the sampler is a continuous denoising procedure enabled by DSL adaptation.

\paragraph{Stochastic term and coordinate choice.} The DSL localization SDE~\cite{dsl2} is natively written in the $z$-space coordinate with an Euler discretization. We instead integrate in the normalized coordinate $y = z/\gamma$, which keeps the state magnitude bounded as $\gamma$ grows and is more stable for long LLaDA sequences. The denoising direction is algebraically the same; only the coordinate and solver order differ. In $y$-space the noise injection acquires an additional $1/\gamma$ factor: the stochastic term at each Heun step is $\frac{1}{2}\bigl(\frac{\eta}{\gamma_s}+\frac{\eta}{\gamma_{s+1}}\bigr)\sqrt{|\Delta\gamma|}\,\xi$ (Algorithm~\ref{alg:inference}). This injection improves sample diversity and prevents mode collapse at low NFE. Setting $\eta = 0$ yields a deterministic denoising trajectory. We find $\eta > 0$ beneficial at low NFE and negligible at high NFE (Table~\ref{tab:sde_impl_sensitivity}).

\begin{algorithm}[ht!]
\caption{DSL Training (one step)}
\label{alg:training}
\begin{algorithmic}[1]
\Require Token sequence $x_{1:L}$, noise embeddings $\{e_v\}$, converter params $W_{\text{token}}, b, \beta$, backbone $f_\theta$
\For{each position $i = 1, \dots, L$}
  \State $\gamma_i \sim 0.9 \cdot \text{LogNormal (per-seq)} + 0.1 \cdot \text{ROAR (per-tok)}$ \Comment{SNR}
  \State $\epsilon_i \sim \mathcal{N}(0, I_d)$ \Comment{sample noise}
  \State $z_i = \gamma_i \cdot e_{x_i} + \sqrt{\gamma_i} \cdot \epsilon_i$ \Comment{noisy embedding}
  \State $\pi_i = \text{softmax}(\beta \cdot z_i^\top W_{\text{token}}^\top + b^\top)$ \Comment{soft token dist.}
  \State $h_i = \pi_i \cdot W_{\text{backbone}}$ \Comment{token mixture}
\EndFor
\State Logits $= f_\theta(h_{1:L})$ \Comment{backbone forward}
\State Loss $= \text{CrossEntropy}(\text{Logits}, x_{1:L})$
\State Update $\theta, W_{\text{token}}, b, \beta$ via gradient descent
\end{algorithmic}
\end{algorithm}

\begin{algorithm}[ht!]
\caption{SDE Inference (Heun solver)}
\label{alg:inference}
\begin{algorithmic}[1]
\Require Context $c_{1:m}$, response length $L$, steps $N/2$, converter params $W_{\text{token}}, b, \beta_{\text{inf}}$, backbone $f_\theta$, noise scale $\eta$, SNR schedule $\gamma_0 < \gamma_1 < \cdots < \gamma_{N/2}$
\State $h_i^{\text{ctx}} = W_{\text{backbone}}[c_i]$ for context positions $i \leq m$ \Comment{clamp context}
\State Sample $\tilde{y}_i \sim \mathcal{N}(0, I_d)$; set $y_i^{(0)} = \tilde{y}_i / \|\tilde{y}_i\|$ for response positions $i = m{+}1, \dots, m{+}L$
\For{$s = 0, \dots, N/2 - 1$}
  \State \textit{// Predict}
  \State $z_i = \gamma_s \cdot y_i^{(s)}$ \Comment{construct converter input}
  \State $h_i = \text{softmax}(\beta_{\text{inf}} \cdot z_i^{\top} W_{\text{token}}^\top + b^\top)\, W_{\text{backbone}}$ for response positions \Comment{convert}
  \State Logits $= f_\theta([h^{\text{ctx}}; h]_{1:m+L})$ \Comment{backbone forward (full sequence)}
  \State $\hat{e}_i^{(s)} = \sum_{v \in \text{top-}M} \tilde{p}(v) \cdot e_v$ \Comment{denoiser from Logits}
  \State $v_i^{(s)} = (\hat{e}_i^{(s)} - y_i^{(s)}) / \gamma_s$ \Comment{$\gamma$-time vector field}
  \State $\tilde{y}_i^{(s+1)} = y_i^{(s)} + \Delta\gamma_s \cdot v_i^{(s)}$ \Comment{Euler predict}
  \State \textit{// Correct}
  \State $\tilde{z}_i = \gamma_{s+1} \cdot \tilde{y}_i^{(s+1)}$; $\tilde{h}_i = \text{softmax}(\beta_{\text{inf}} \cdot \tilde{z}_i^{\top} W_{\text{token}}^\top + b^\top)\, W_{\text{backbone}}$
  \State $\tilde{\text{Logits}} = f_\theta([h^{\text{ctx}}; \tilde{h}]_{1:m+L})$; compute $\hat{e}_i^{(s+1)}$, $v_i^{(s+1)} = (\hat{e}_i^{(s+1)} - \tilde{y}_i^{(s+1)}) / \gamma_{s+1}$
  \State $\xi_i \sim \mathcal{N}(0,I_d)$
  \State $y_i^{(s+1)} = y_i^{(s)} + \frac{\Delta\gamma_s}{2}\bigl(v_i^{(s)} + v_i^{(s+1)}\bigr) + \frac{1}{2}\!\left(\frac{\eta}{\gamma_s}+\frac{\eta}{\gamma_{s+1}}\right)\sqrt{|\Delta\gamma_s|}\,\xi_i$ \Comment{Heun update}
\EndFor
\State $z_i = \gamma_{N/2} \cdot y_i^{(N/2)}$; $h_i = \text{softmax}(\beta_{\text{inf}} \cdot z_i^{\top} W_{\text{token}}^\top + b^\top)\, W_{\text{backbone}}$ \Comment{final convert}
\State Logits $= f_\theta([h^{\text{ctx}}; h]_{1:m+L})$ \Comment{backbone forward}
\State $\hat{x}_i = \arg\max_v \text{Logits}_i$ for response positions \Comment{decode}
\State Truncate at first EOS token
\State \Return $\hat{x}_{m+1:m+L}$
\end{algorithmic}
\end{algorithm}

\section{Experimental Details}
\label{app:details}

\subsection{Training}

All models start from LLaDA-8B-Instruct~\cite{llada}, continued-pretrained on FineWeb-Edu (sample-10BT, streaming) for 1{,}000 steps. The primary DSL-LLaDA configuration uses $\beta{=}1$ with random unit-norm noise embeddings (the primary setting for continuous SDE sampling). Appendix-only ablation runs use $\beta\in\{0.3,1,2,3,4,5\}$, semantic noise embeddings (3{,}000 steps), or ROAR-only SNR.

\begin{table}[ht!]
\centering
\caption{Training hyperparameters.}
\label{tab:hparams}
\footnotesize
\begin{tabular}{@{}>{\raggedright\arraybackslash}p{0.26\linewidth}@{\quad}>{\raggedright\arraybackslash}p{0.62\linewidth}@{}}
\toprule
Hardware & 8$\times$A100-80GB \\
Parallelism & DeepSpeed ZeRO-2, bf16 mixed precision \\
Optimizer & AdamW ($\beta_1{=}0.9$, $\beta_2{=}0.999$, wd$=$0.1) \\
Learning rate & $2 \times 10^{-5}$, constant w/ warmup 100 steps \\
Batch size & 32 (per-GPU$=$1, grad.\ accum.$=$4, $\times$8 GPU) \\
Sequence length & 2{,}048 tokens \\
Gradient clipping & 1.0 \\
\midrule
\multicolumn{2}{@{}l}{\textit{DSL-specific (primary, $\beta{=}1$, random embed)}} \\
Noise embedding & $d{=}100$, unit-norm (iid $\mathcal{N}(0,I)$, row-normalized), trainable (re-normalized each forward pass) \\
Converter $\beta$ init & 1.0 (learnable, $\beta{=}1$ at convergence) \\
Converter softmax & float32 \\
Converter LR & $25\times$ backbone ($5 \times 10^{-4}$) \\
SNR distribution & 90\% per-sequence LogNormal($\mu{=}1.69$, $\sigma{=}0.9$, clip 40) + 10\% smoothed ROAR (clear $\sim U[80,100]$, unknown $\sim U[0,1)$) \\
SNR granularity & Per-sequence (LogNormal) / per-token (ROAR) \\
\midrule
\multicolumn{2}{@{}l}{\textit{Appendix ablations (Appendix~\ref{app:design_space})}} \\
Other $\beta$ & 0.3, 2, 3, 4, 5 (all random embed, 1K steps) \\
Semantic embed & $d{=}100$, AE+contrastive from wte; $\beta{=}0.5$, 3K steps \\
\bottomrule
\end{tabular}
\end{table}

\subsection{SDE Inference Hyperparameters}
\label{app:sde_impl}

Table~\ref{tab:sde_inference_hparams} lists the inference settings used by DSL-LLaDA-SDE. The main experiments use the sensitive SNR schedule and $M{=}512$ for the denoiser expectation. The stochastic noise scale is tied to the NFE budget: higher noise helps very small NFE runs avoid early collapse, while NFE$=64$ uses only a small perturbation.

\begin{table}[ht!]
\centering
\caption{Default SDE inference hyperparameters. The Heun solver uses $N/2$ solver steps and two model evaluations per step, for $N$ total forward evaluations.}
\label{tab:sde_inference_hparams}
\resizebox{\columnwidth}{!}{%
\begin{tabular}{@{}ll@{}}
\toprule
Parameter & Value \\
\midrule
Solver & Heun predictor-corrector \\
NFE & 8, 16, 32, 64 (128 in generation appendix) \\
SNR range & 0.01 to 100 \\
SNR schedule & 5\% steps: 0.01--7; 90\%: 7--74; 5\%: 74--100 \\
Stochastic scale $\eta$ & 0.10 (NFE$\leq$8), 0.05 (NFE$\leq$16), 0.01 (NFE$\leq$32), 0.005 otherwise \\
Denoiser top-$M$ & 512 \\
Inference converter multiplier & 2.0 unless otherwise stated \\
Decode & argmax over backbone token logits \\
\bottomrule
\end{tabular}}
\end{table}

\subsection{ReMDM Inference Hyperparameters}
\label{app:rmdm_impl}

Table~\ref{tab:rmdm_hparams} lists the inference settings for the LLaDA ReMDM baseline. We use the ReMDM-conf-loop variant~\cite{remdm}: the denoising trajectory is divided into three phases by the loop window $[t_{\text{off}}, t_{\text{on}}]$. In the early phase ($t > t_{\text{on}}$) and late phase ($t \leq t_{\text{off}}$), standard unmasking is applied without remasking. In the middle loop phase ($t_{\text{off}} < t \leq t_{\text{on}}$), the noise schedule is held constant at $\alpha = \alpha(t_{\text{on}})$: for each already-unmasked token, an uncertainty score $1 - \text{confidence}$ is computed and normalized via softmax across all unmasked positions, yielding a per-token probabilistic remask probability. Low-confidence tokens are stochastically re-masked and re-predicted while high-confidence tokens are retained. There is no fixed confidence threshold.

\begin{table}[ht!]
\centering
\caption{LLaDA ReMDM inference hyperparameters (ReMDM-conf-loop).}
\label{tab:rmdm_hparams}
\footnotesize
\resizebox{\columnwidth}{!}{%
\begin{tabular}{@{}ll@{}}
\toprule
Parameter & Value \\
\midrule
Weights & LLaDA-8B-Instruct (original, no extra training) \\
Method flag & \texttt{--method rmdm} (internal: \texttt{remdm\_conf}) \\
Loop window on ($t_{\text{on}}$) & 0.55 \\
Loop window off ($t_{\text{off}}$) & 0.05 \\
Loop alpha ($\alpha_{\text{loop}}$) & 0.9 \\
Temperature & 0.0 \\
CFG & 0.0 \\
EOS suppression & not enabled \\
Block length & 256 (clamped to gen length if smaller) \\
NFE & 8, 16, 32, 64 \\
\midrule
\multicolumn{2}{@{}l}{\textit{Summarization generation lengths}} \\
XSum & 128 \\
CNN/DM, PubMed, arXiv & 256 \\
Long-form & 512, 1{,}024 \\
\midrule
Samples & 1{,}000 per benchmark, seed 42 \\
\bottomrule
\end{tabular}}
\end{table}

\paragraph{Methods compared.} To avoid naming ambiguity across tables, we fix the following shorthand throughout the paper. Baselines use the \textbf{LLaDA-...} prefix; our DSL-trained model uses the \textbf{DSL-LLaDA-...} prefix; the suffix indicates inference mode.
\begin{itemize}\setlength\itemsep{1pt}
\item \textbf{LLaDA}: LLaDA-8B-Instruct~\cite{llada} with default discrete remasking inference.
\item \textbf{LLaDA+EOS}: LLaDA with the EOS logit forced to $-\infty$ during sampling (prevents EOS cascade).
\item \textbf{LLaDA+EOS+Block} (denoted \textbf{LLaDA$^\star$} in the main text): LLaDA+EOS combined with block remasking (block$=$32 tokens). This is our strongest length-controlled discrete heuristic among the settings we tested: EOS suppression prevents premature truncation, while block remasking unmasks 32 contiguous tokens per step to encourage local coherence. We use this as an engineered discrete comparison throughout the paper.
\item \textbf{LLaDA+Block}: LLaDA with block remasking only (no EOS suppression).
\item \textbf{MDM-CPT}: our \emph{discrete control}: LLaDA-8B continue-pretrained on the same data and compute as DSL, but with standard MDLM random-time binary masking (per-sequence $t\sim U(0,1)$, mask probability $p=(1{-}\epsilon)t+\epsilon$, $\epsilon{=}10^{-3}$, loss on masked positions weighted by $1/p$). Isolates the effect of continuous noise vs.\ continued pretraining.
\item \textbf{XDLM}~\cite{xdlm}: a re-implementation of the XDLM/XDM forward process as continued pretraining of LLaDA (same data, steps, batch size, and learning rate as DSL-LLaDA and MDM-CPT). At each training step, a random diffusion time $t\sim U(0,1)$ determines the keep probability $\alpha_t = 1-(1-\epsilon)t-\epsilon$ ($\epsilon{=}10^{-3}$); among non-kept positions, $k_1{=}10\%$ of the corruption mass is random-vocabulary replacement and the remainder is \texttt{[MASK]}. We also evaluate the publicly released XDLM Base checkpoint with its native xDM sampler in Appendix~\ref{app:xdlm_public}; its weaker summarization scores reflect the lack of instruction tuning rather than a sampler limitation.
\item \textbf{LLaDA ReMDM}: LLaDA-8B-Instruct with ReMDM confidence-loop remasking~\cite{remdm} (\texttt{remdm\_conf}). This baseline changes inference only and does not update the original LLaDA weights; it uses the same NFE budget, generation length, prompt format, temperature (0.0), and CFG (0.0) as the corresponding LLaDA discrete baselines. Hyperparameters are listed in Table~\ref{tab:rmdm_hparams}.
\item \textbf{DSL-LLaDA-Remask}: our DSL-trained model using discrete remasking at inference (converter sharpness $\beta{=}1$ unless otherwise noted).
\item \textbf{DSL-LLaDA-Remask+EOS} / \textbf{DSL-LLaDA-Remask+Block}: DSL-LLaDA-Remask combined with EOS suppression / block remasking (appendix only, to show the DSL model's discrete-inference failures are not EOS artifacts).
\item \textbf{DSL-LLaDA-SDE} (ours, primary): our DSL-trained model with continuous SDE inference in embedding space (Heun solver, $\beta{=}1$ training configuration).
\item \textbf{$\beta$ / embedding ablations}: alternate DSL configurations (large-$\beta$ random embedding, small-$\beta$ semantic embedding, ROAR-only SNR) explored in Appendix~\ref{app:design_space}; these support discrete remasking but degrade the pure SDE solver, so DSL-LLaDA-SDE in the main text always refers to the $\beta{=}1$ random-embedding configuration.
\end{itemize}
Column headers in dense tables (e.g., Table~\ref{tab:gen_full}) may use abbreviations ``SDE''$=$DSL-LLaDA-SDE, ``Rmk''$=$DSL-LLaDA-Remask, ``Orig''$=$LLaDA, ``MDM''$=$MDM-CPT.

\subsection{Evaluation Settings}

All evaluations use seed 42 and bf16 inference. The evaluation corpus comprises 100 texts from WikiText-103 (test split, 20--80 words each), cached as \texttt{texts\_100.json}. Corruption, calibration, and generation evaluations share this fixed corpus.

\paragraph{Error correction.} 100 texts. Random corruption: replace $r$\% of positions with random token IDs $\in [100, 126000)$. Single forward pass. Metrics: fix rate (corrupted$\to$correct) and clean preserved (uncorrupted$\to$unchanged). The primary $\beta{=}1$ DSL-LLaDA model targets random corruption only (the $\beta{=}1$ converter cannot reliably distinguish semantic neighbors); semantic-corruption results for higher-$\beta$ ablations appear in Appendix~\ref{app:design_space}.

\paragraph{Surface corruption.} Tokenizer-aligned spelling, OCR, and homophone corruptions (26 corrupted tokens across 26 cases, 324 clean tokens). Construction details and the expanded 1{,}000-text noisy-edit evaluation appear in Appendix~\ref{app:design_space}.

\paragraph{Context robustness.} 100 texts, 30\% masked, corrupt $c$\% of unmasked positions. Metric: top-1 accuracy on masked positions.

\paragraph{Generation quality.} 200 prompts, chat template, gen=256, seed 42. GenPPL: GPT-2 Large perplexity truncated to 128 words. D2 (Distinct-2): fraction of unique bigrams across all generated texts (higher = more diverse). Rep: the fraction of adjacent generated word pairs with identical whitespace-delimited forms, a local unigram analogue of the $n$-gram repetition diagnostics commonly used to study neural text degeneration~\citep{holtzman2020curious,welleck2020neural} (lower = less degenerate). Len: average word count of generated text.

\paragraph{Summarization dataset sourcing.} XSum~\cite{xsum} is loaded from the \texttt{EdinburghNLP/xsum} HuggingFace repository. CNN/DailyMail~\cite{cnndm} uses \texttt{cnn\_dailymail} v3.0.0. For PubMed and arXiv~\cite{pubmed_summ} the original \texttt{scientific\_papers} dataset ships as a deprecated dataset-script and was replaced with \texttt{ccdv/pubmed-summarization} and \texttt{ccdv/arxiv-summarization}, which pre-parse the same papers into parquet; reference summaries are slightly longer than in the original (PubMed 205 vs.\ 146 words, arXiv 163 vs.\ 129). Source documents longer than 1{,}500 words are truncated. All splits are the official \texttt{test} split and we take 1{,}000 examples uniformly at random with seed 42.

\paragraph{Metrics.} We report ROUGE-1/2/L F1 against reference summaries following standard summarization evaluation. Per-sample word counts (Len), degenerate-ending rate (Degen\%), and automated failure-mode tagging are computed post-hoc from the generated text.

\paragraph{BERTScore.} Appendix~\ref{app:sum_bertscore} reports BERTScore (roberta-large, rescaled with baseline) for the NFE=64 runs. BERTScore is monotone-consistent with ROUGE: SDE leads on XSum and CNN/DM, LLaDA leads on PubMed and arXiv.

\paragraph{LLM-as-judge.} We use Azure OpenAI GPT-5.4 as a pairwise judge, with A/B positions randomized per sample to eliminate order bias. Each call returns four 1--5 scores (factuality, coverage, fluency, conciseness) plus an overall preference. About 0.3\% of samples are filtered by Azure's content policy (news articles containing sensitive topics); these contribute ``no judgment'' rather than a biased score.

\section{Converter Output Examples}
\label{app:converter}

\begin{table}[ht!]
\centering
\caption{Converter output for ``Paris'' and digit ``7'' at each SNR (20 random seeds). Three phases: unknown (mask-like), unreliable (random), clear (gold + neighbors).}
\label{tab:converter}
\resizebox{\columnwidth}{!}{%
\begin{tabular}{rcccc cccc}
\toprule
& \multicolumn{4}{c}{``Paris''} & \multicolumn{4}{c}{``7''} \\
\cmidrule(lr){2-5}\cmidrule(lr){6-9}
SNR & top-1 & gold cos & hit & & top-1 & top-2 & top-3 & hit \\
\midrule
1  & [MASK] & +.15 & 0 & & --- & --- & --- & 0 \\
10 & \textit{random} & +.16 & 1 & & \textit{random} & --- & --- & 0 \\
25 & \best{Paris}(9) & +.43 & 9 & & \best{7}(14) & \best{6} & \best{8} & 14 \\
40 & \best{Paris}(15) & +.75 & 15 & & \best{7}(19) & \best{6} & \best{8} & 19 \\
50 & \best{Paris}(20) & +.96 & 20 & & \best{7}(20) & \best{6} & \best{8} & 20 \\
\bottomrule
\end{tabular}}
\end{table}

\section{Ablations and Sensitivity Analysis}
\label{app:ablations_all}

This section consolidates all ablation studies and sensitivity analyses. We first examine the converter design space and training-time choices (\S\ref{app:design_space}), then present controlled ablations on error correction (\S\ref{app:ablations}), verify that DSL training preserves the original mask-filling capability (\S\ref{app:mask_calibration}), and finally sweep inference-time hyperparameters (\S\ref{app:sde_sensitivity}).

\subsection{Converter Design Space}
\label{app:design_space}

Two knobs determine what the backbone sees at low SNR: $\beta$ (logit sharpness) and noise embedding geometry.

\paragraph{Illustrative example.}
Consider a corrupted sentence where two tokens have been replaced with random words:
\begin{center}
\small
\resizebox{\columnwidth}{!}{%
\begin{tabular}{@{}r@{\;:\;}l@{}}
Ground truth & The \textbf{cat} is \textbf{on} the laptop. \\
Corrupted    & The \textcolor{red}{lemon} is \textcolor{red}{tree} the laptop. \\
XDLM         & The \textbf{cat} is \textbf{on} the \textit{\textcolor{red}{fridge}}. \quad {\scriptsize [changes an uncorrupted token]} \\
DSL           & The \textbf{cat} is \textit{\textcolor{blue}{under}} the laptop. \quad {\scriptsize [incorrect change to corrupted token]} \\
\end{tabular}}
\end{center}
XDLM fixes both corrupted tokens but also overwrites the clean token ``laptop'' (low Clean). DSL preserves all clean tokens but replaces ``on'' with a plausible but incorrect alternative (high Clean, imperfect Fix). This tradeoff between fix rate and clean-token preservation is precisely what the Fix/Clean metrics in Table~\ref{tab:corruption} capture.

\paragraph{Random embedding + large $\beta$.} The primary configuration uses random embedding with $\beta{=}1$; ablations use $\beta{=}2{-}5$. At low SNR the law of large numbers averages 126K near-orthogonal token embeddings into a single MASK-like point, creating a binary ``unknown or clear'' signal. The binary signal trains the backbone to detect random corruptions but not semantically plausible errors.

\paragraph{Semantic embedding + small $\beta$.} With e.g.\ $\beta{=}0.5$ and AE+contrastive embeddings, semantic neighborhoods emerge gradually as SNR increases, creating a graded reliability signal. The backbone learns to detect contextually implausible tokens even when they are embedding neighbors.

\paragraph{ROAR-only SNR.} Dropping the LogNormal mixture, each position independently receives either low SNR ${\sim}\,U[0,1)$ or high SNR ${\sim}\,U[80,100]$, creating explicit within-sequence context/target structure with soft corruption. This produces a model that detects both random and semantic errors at $\beta{=}2$ (Table~\ref{tab:ablations}: 64\% + 77\%).

Large $\beta$ + semantic embedding fails because semantic neighbors receive deterministic logit boosts at low SNR, breaking the law of large numbers (training loss plateaus at 4.6 vs.\ 2.0 for the random-embedding configuration).

\begin{table}[ht!]
\centering
\caption{Realistic selective-correction probes (Fix/Clean, \%). Surface = 26 aligned typo/OCR/homophone cases; entity/number and semantic-neighbor cases are manually constructed plausible substitutions. The primary $\beta{=}1$ model is highly clean-preserving and fixes some surface errors, but semantic or factual substitutions require separate correction-oriented variants.}
\label{tab:realistic_correction}
\resizebox{\columnwidth}{!}{%
\begin{tabular}{lcccc}
\toprule
\textbf{Model} & \textbf{Random@30} & \textbf{Surface} & \textbf{Entity/Num.} & \textbf{Semantic} \\
\midrule
LLaDA        & 5.7 / 89.4  & 26.9 / 78.8 & 15.1 / 91.2 & 10.2 / 91.3 \\
MDM-CPT      & 10.3 / 90.6 & 26.9 / 88.5 & 9.1 / 92.9  & 6.8 / 93.4 \\
XDLM         & \best{55.7} / 71.1 & 46.2 / 87.2 & \best{30.3} / 92.0 & \best{29.7} / 90.6 \\
\ours{} ($\beta{=}1$) & 48.5 / \best{98.7} & 46.2 / 98.4 & 3.0 / 99.0 & 3.4 / 98.0 \\
\ours{} ($\beta{=}2$) & 54.3 / 98.4 & 46.2 / \best{99.9} & 1.5 / \best{99.4} & 1.7 / 98.4 \\
\ours{} (semantic) & 20.2 / 98.0 & 46.2 / 98.9 & 24.2 / 98.6 & 6.8 / \best{98.5} \\
\bottomrule
\end{tabular}}
\end{table}

\paragraph{Expanded evaluation (1{,}000 texts).}
The main-text correction table uses 100 texts for consistency with the development probes. Table~\ref{tab:correction_1k} reports the same random-token evaluation on 1{,}000 WikiText texts alongside an automatic noisy-edit benchmark. Noisy edits are tokenizer-aligned spelling corruptions (single-character deletion or adjacent swap), OCR-like confusions (e.g., \texttt{rn}$\to$\texttt{m}), and homophone substitutions, retaining only replacements that map to a single token. The generator yields 770 valid texts with 2{,}659 corrupted tokens.

\begin{table}[ht!]
\centering
\caption{Random-token and noisy-edit correction on 1{,}000 WikiText texts (single forward pass). Random columns report Fix/Clean at 10\%, 30\%, and 50\% corruption rates. Noisy-edit reports Fix/Clean over 770 valid texts (2{,}659 corrupted tokens). DSL-LLaDA preserves $>$93\% clean tokens under random corruption and $>$99\% under noisy edits.}
\label{tab:correction_1k}
\resizebox{\columnwidth}{!}{%
\begin{tabular}{l cc cc cc cc}
\toprule
& \multicolumn{2}{c}{\textbf{Random @10\%}} & \multicolumn{2}{c}{\textbf{Random @30\%}} & \multicolumn{2}{c}{\textbf{Random @50\%}} & \multicolumn{2}{c}{\textbf{Noisy-Edit}} \\
\cmidrule(lr){2-3}\cmidrule(lr){4-5}\cmidrule(lr){6-7}\cmidrule(lr){8-9}
\textbf{Model} & Fix & Clean & Fix & Clean & Fix & Clean & Fix & Clean \\
\midrule
LLaDA              & 15.0 & 87.7 & 7.5 & 85.5 & 1.6 & 81.3 & 34.7 & 92.9 \\
XDLM               & \best{69.3} & 80.8 & \best{59.7} & 72.1 & \best{41.9} & 54.1 & \best{58.5} & 89.4 \\
\ours{}\,($\beta{=}1$) & 66.1 & \best{95.0} & 55.1 & \best{94.4} & 36.5 & \best{93.2} & 49.4 & \best{99.1} \\
\bottomrule
\end{tabular}}
\end{table}

The expanded random-token results are consistent with the 100-text development set: DSL-LLaDA closes most of the fix-rate gap with XDLM while preserving substantially more clean tokens. On noisy edits, DSL-LLaDA's clean-token preservation rises to 99.1\%, reflecting the model's conservatism on subtle corruptions that resemble valid tokens. XDLM achieves a higher fix rate (58.5\% vs.\ 49.4\%) but overcorrects 10.6\% of clean tokens.

\subsection{Training Ablations}
\label{app:ablations}

SDE inference operates in a continuous embedding space, which requires a model trained to denoise continuous inputs; standard masked diffusion models such as LLaDA and MDM-CPT only see discrete tokens and therefore cannot be evaluated under SDE. The ablations below instead isolate design choices \emph{within} the DSL framework. Table~\ref{tab:ablations} reports two ablations: (i) converter sharpness $\beta$ with random embedding (top group), and (ii) embedding type, SNR distribution, and architecture at $\beta{=}2$ (bottom group). All runs use the primary mixed SNR scheme (per-sequence LogNormal + per-token smoothed ROAR) unless otherwise noted, with $d{=}100$ and 1{,}000 training steps (Sem $\beta{=}0.5$ uses 3K).

\begin{table}[ht!]
\centering
\caption{Ablations on error correction. Top: $\beta$ sweep (random embedding). Bottom: embedding/architecture variants at $\beta{=}2$ (except Sem $\beta{=}0.5$, 3K steps). Random fix increases monotonically with $\beta$; semantic fix is governed by embedding geometry and SNR distribution.}
\label{tab:ablations}
\resizebox{\columnwidth}{!}{%
\begin{tabular}{l cccc}
\toprule
Variant & Rand Fix & Sem Fix & Clean & ECE \\
\midrule
\multicolumn{5}{l}{\textit{$\beta$ sweep (random embedding, 1K steps)}} \\[2pt]
$\beta{=}0.3$           &  8.5       & 23.8        & 99.1        & 0.043        \\
$\beta{=}1.0$           & 64.3       & \best{32.2} & 99.0        & 0.026        \\
$\beta{=}2.0$ (base)    & 64.8       & 19.5        & 99.0        & \best{0.023} \\
$\beta{=}3.0$           & 65.5       & 18.4        & 98.5        & 0.025        \\
$\beta{=}4.0$           & 67.0       & 20.0        & 98.7        & 0.029        \\
$\beta{=}5.0$           & \best{69.2}& 22.5        & 98.6        & 0.032        \\
\midrule
\multicolumn{5}{l}{\textit{Embedding / architecture (relative to $\beta{=}2$ base)}} \\[2pt]
\quad + frozen embed    & 63.2       & 21.6        & \best{99.2} & 0.031        \\
\quad + ROAR 100\%      & 64.4       & \best{77.4} & 98.5        & \best{0.022} \\
\quad + highpass SNR    & 37.5       & 34.7        & 98.5        & 0.028        \\
\quad + residual        & 15.1       & 29.3        & 99.1        & 0.031        \\
Sem embed $\beta{=}2$ (1K)   & 65.3 & 70.3        & 97.3        & 0.037        \\
Sem embed $\beta{=}0.5$ (3K) & 42.0 & 66.7        & 98.8        & 0.038        \\
\bottomrule
\end{tabular}}
\end{table}

\paragraph{$\beta$ sweep.} $\beta$ controls logit sharpness and thus the SNR threshold at which the gold token emerges. Higher $\beta$ sharpens the MASK-to-token transition, giving the backbone a clearer signal about ``obviously wrong'' tokens, explaining the monotonic increase in random fix rate (8.5\%$\to$69.2\%). Semantic fix shows no consistent $\beta$ dependence; ECE is best at $\beta{=}2$.

\paragraph{Frozen embedding} performs nearly identically to the trained baseline, confirming DSL's value comes from the converter and continuous noise rather than learned embedding geometry.

\paragraph{ROAR 100\%} (pure per-position binary SNR, no LogNormal) achieves the highest semantic fix (77.4\%) while maintaining the random-embedding $\beta{=}2$ random fix rate (64.4\%). Each position independently receives either low SNR (soft noise) or high SNR (clear token), creating explicit within-sequence context/target structure with soft corruption.

\paragraph{Highpass and residual} both degrade random fix substantially, suggesting that the standard SNR distribution and the direct converter-to-backbone path are important.

\paragraph{Semantic embed + $\beta{=}2$} achieves strong semantic fix (70.3\%) at only 1K steps despite its high training loss (4.6 vs.\ 2.0 for the random-embedding setting), suggesting room for further improvement with longer training.

\subsection{Mask Prediction and Calibration}
\label{app:mask_calibration}

\begin{table}[ht!]
\centering
\caption{Mask prediction accuracy and ECE at 30\%/50\%/70\% masking. All models preserve LLaDA's mask-filling ability; ECE improves for most DSL variants.}
\label{tab:mask_full}
\resizebox{\columnwidth}{!}{%
\begin{tabular}{l cccccc}
\toprule
& \multicolumn{2}{c}{30\%} & \multicolumn{2}{c}{50\%} & \multicolumn{2}{c}{70\%} \\
\cmidrule(lr){2-3}\cmidrule(lr){4-5}\cmidrule(lr){6-7}
Model & Acc & ECE & Acc & ECE & Acc & ECE \\
\midrule
LLaDA & 65.1 & .039 & 52.2 & .037 & 30.4 & .049 \\
MDM-CPT & 65.2 & \best{.020} & 51.3 & \best{.016} & 29.0 & .030 \\
XDLM & 65.3 & .026 & 51.3 & .020 & 29.6 & .029 \\
DSL ($\beta{=}1$, primary) & \best{67.0} & .026 & 52.2 & .032 & 30.2 & .030 \\
DSL ($\beta{=}2$, ablation) & 66.6 & .023 & \best{52.5} & .021 & \best{31.2} & \best{.021} \\
DSL (ROAR-only, $\beta{=}2$) & 65.3 & .022 & 51.6 & .019 & 29.8 & .025 \\
DSL (sem embed, $\beta{=}0.5$) & 64.8 & .038 & 50.5 & .032 & 29.4 & .037 \\
DSL (frozen embed, $\beta{=}2$) & 66.9 & .031 & 52.3 & .017 & 30.6 & .027 \\
\bottomrule
\end{tabular}}
\end{table}

All DSL variants maintain mask-prediction accuracy within 2\% of LLaDA. ECE improves consistently across the board: the primary $\beta{=}1$ DSL-LLaDA achieves accuracy 67.0\% at 30\% masking (best in the table) while reducing ECE from 0.039 (LLaDA) to 0.026; the $\beta{=}2$ ablation reaches the lowest ECE among DSL variants (0.023). The semantic-embedding variant ($\beta{=}0.5$) lands closer to LLaDA's calibration (0.038), reflecting its weaker random-token detection.

\subsection{SDE Inference Sensitivity}
\label{app:sde_sensitivity}

We run a low-cost sensitivity sweep on the primary $\beta{=}1$ checkpoint to verify that the SDE sampler does not depend on a single narrow hyperparameter setting. Open-ended generation uses 100 prompts at NFE=64 and gen=256; summarization uses 100 XSum examples at NFE=64. Table~\ref{tab:sde_impl_sensitivity} reports the main trends.

\begin{table}[ht!]
\centering
\caption{SDE implementation sensitivity. Open-ended metrics are GenPPL / D2 / Rep / Len; XSum reports R-1/R-2/R-L. Top-$M$ and inference $\beta$ have broad usable ranges, while the sensitive SNR schedule is important.}
\label{tab:sde_impl_sensitivity}
\resizebox{\columnwidth}{!}{%
\begin{tabular}{ll cccc ccc}
\toprule
Sweep & Setting & GenPPL & D2 & Rep & Len & R-1 & R-2 & R-L \\
\midrule
\multirow{4}{*}{top-$M$}
& 128  & 5.19 & .320 & .030 & 181.6 & 33.40 & 11.84 & 26.16 \\
& 256  & 4.88 & .315 & .025 & 183.6 & 32.68 & 11.75 & 25.22 \\
& 512  & 4.77 & .321 & .031 & 182.3 & 33.22 & 12.04 & 25.99 \\
& 1024 & 4.74 & .313 & .035 & 182.1 & 32.69 & 11.63 & 25.70 \\
\midrule
\multirow{6}{*}{schedule / $\eta$}
& sensitive, 0.000 & 4.68 & .314 & .059 & 183.2 & 32.83 & 12.06 & 25.74 \\
& sensitive, 0.005 & 4.77 & .321 & .031 & 182.3 & 33.22 & 12.04 & 25.99 \\
& sensitive, 0.010 & 4.98 & .326 & .024 & 181.8 & 33.06 & 12.15 & 26.01 \\
& uniform, 0.000   & 6.08 & .204 & .223 & 152.1 & 31.61 & 11.18 & 25.00 \\
& uniform, 0.005   & 6.66 & .216 & .188 & 153.0 & 31.89 & 10.69 & 25.07 \\
& uniform, 0.010   & 6.73 & .224 & .183 & 153.2 & 31.82 & 10.68 & 25.17 \\
\midrule
\multirow{3}{*}{inference $\beta$}
& 1.5 & 5.07 & .297 & .074 & 173.6 & 32.74 & 11.69 & 25.90 \\
& 2.0 & 4.63 & .321 & .028 & 183.9 & 32.63 & 11.77 & 25.24 \\
& 2.5 & 5.28 & .345 & .003 & 199.2 & 32.09 & 11.43 & 25.16 \\
\bottomrule
\end{tabular}}
\end{table}

Two conclusions are useful for reproducibility: (1)~top-$M$ is not a sensitive hyperparameter, as values from 128 to 1024 keep both open-ended and summarization metrics in the same range, and (2)~the SNR schedule matters. Replacing it with a uniform log-SNR schedule increases open-ended repetition by roughly 6--9$\times$ and shortens generations. The stochastic scale and inference $\beta$ show moderate sensitivity rather than sharp failure points: $\eta{=}0.005$ balances repetition and XSum quality at NFE=64, and $\beta{=}2.0$ gives the best overall profile among the three tested inference multipliers.

\section{Summarization Analysis Details}
\label{app:sum_analysis}

We provide the full per-sample analysis underlying Table~\ref{tab:sum_quality} in the main text. All numbers are at NFE=64 on the same 1{,}000-sample test subsets; the LLM-judge evaluation uses 100 prompt-randomized samples per dataset.

\subsection{LLaDA+EOS+Block: a third baseline}
\label{app:block_baseline}

Table~\ref{tab:summarization_block_full} adds LLaDA+EOS+Block to the comparison at every NFE. Two findings: (1)~block-remasking underperforms at low NFE (R-1 below 15 at NFE$\leq$16 on all benchmarks), and (2)~at NFE=64 it introduces degenerate dot/comma-spam endings in 7--25\% of short-reference outputs (Appendix~\ref{app:sum_failures}). Table~\ref{tab:judge_axes_block} shows SDE wins by wider margins against LLaDA$^\star$ than against vanilla LLaDA (mean 59.8\% vs.\ 46.7\% at NFE=64), driven by fluency.

\begin{table}[ht!]
\centering
\caption{Four-way comparison at NFE$\in\{8,16,32,64\}$ (1{,}000 samples, ROUGE-1 F1). Bold marks per-cell maxima. LLaDA+EOS+Block underperforms vanilla LLaDA on the long-reference benchmarks at NFE=64 and produces high degenerate-ending rates on the short-reference benchmarks (Appendix~\ref{app:sum_failures}). XDLM Base + xDM is the publicly released XDLM checkpoint with its native xDM sampler ($k_1{=}0.1$, plain prompt); it lacks instruction tuning and trails all instruction-tuned methods (see Appendix~\ref{app:xdlm_public}).}
\label{tab:summarization_block_full}
\resizebox{\columnwidth}{!}{%
\begin{tabular}{ll cccc}
\toprule
Dataset (ref) & Method & NFE=8 & NFE=16 & NFE=32 & NFE=64 \\
\midrule
\multirow{4}{*}{XSum (21w)}
& DSL-LLaDA-SDE     & \best{28.4} & \best{30.4} & \best{32.0} & \best{32.4} \\
& LLaDA             & 25.2        & 29.0        & 28.8        & 24.8        \\
& LLaDA+EOS+Block   & 9.9         & 14.1        & 20.2        & 24.1        \\
& XDLM Base + xDM   & 4.5         & 7.6         & 14.4        & 15.4        \\
\midrule
\multirow{4}{*}{CNN/DM (55w)}
& DSL-LLaDA-SDE     & \best{28.1} & \best{33.0} & \best{35.1} & \best{35.8} \\
& LLaDA             & 23.1        & 28.2        & 25.4        & 18.2        \\
& LLaDA+EOS+Block   & 8.6         & 8.3         & 18.7        & 28.4        \\
& XDLM Base + xDM   & 3.3         & 4.6         & 11.0        & 23.0        \\
\midrule
\multirow{4}{*}{PubMed (205w)}
& DSL-LLaDA-SDE     & \best{29.4} & \best{32.2} & \best{36.9} & 39.5        \\
& LLaDA             & 11.9        & 20.2        & 36.5        & \best{42.2} \\
& LLaDA+EOS+Block   & 6.1         & 7.5         & 20.9        & 37.8        \\
& XDLM Base + xDM   & 7.5         & 7.7         & 10.9        & 18.9        \\
\midrule
\multirow{4}{*}{arXiv (163w)}
& DSL-LLaDA-SDE     & \best{27.3} & \best{28.6} & 32.9        & 34.9        \\
& LLaDA             & 10.6        & 16.2        & \best{33.8} & \best{40.8} \\
& LLaDA+EOS+Block   & 7.8         & 6.6         & 15.9        & 34.7        \\
& XDLM Base + xDM   & 6.0         & 6.3         & 8.8         & 14.6        \\
\bottomrule
\end{tabular}}
\end{table}

\begin{table}[ht!]
\centering
\caption{LLM-as-judge per-axis means: DSL-LLaDA-SDE vs.\ LLaDA+EOS+Block (100 samples/dataset, GPT-5.4, 1--5 scale). SDE wins fluency on every benchmark (gains from 0.68 to 1.81 points), helping explain the consistent preference gap over the length-controlled baseline.}
\label{tab:judge_axes_block}
\resizebox{\columnwidth}{!}{%
\begin{tabular}{l rrrr rrrr r}
\toprule
& \multicolumn{4}{c}{DSL-LLaDA-SDE} & \multicolumn{4}{c}{LLaDA+EOS+Block} & \\
\cmidrule(lr){2-5}\cmidrule(lr){6-9}
Dataset & fact & cov & flu & conc & fact & cov & flu & conc & SDE-win\% \\
\midrule
XSum    & \best{3.04} & 2.66 & \best{4.60} & \best{4.60} & 2.71 & \best{3.24} & 2.79 & 2.64 & \best{61.0} \\
CNN/DM  & \best{3.12} & 2.73 & \best{3.14} & \best{3.23} & 2.66 & \best{3.25} & 1.54 & 1.57 & \best{59.1} \\
PubMed  & \best{2.66} & \best{2.85} & \best{2.89} & \best{2.45} & 2.44 & 2.53 & 1.86 & 2.11 & \best{59.0} \\
arXiv   & \best{2.46} & \best{2.59} & \best{2.34} & \best{2.11} & 2.22 & 2.16 & 1.66 & 1.91 & \best{60.0} \\
\bottomrule
\end{tabular}}
\end{table}

\subsection{Tuned discrete baselines}
\label{app:discrete_tuning}

We sweep discrete-remasking hyperparameters (block size, EOS handling, temperature) at NFE=16 to check whether the SDE gains come from undertuned baselines (Table~\ref{tab:discrete_tuning}).

\begin{table}[ht!]
\centering
\caption{Discrete-remasking development sweep at NFE=16. Default full-sequence remasking with mild temperature gives the best ROUGE but short outputs; EOS forcing lengthens outputs but increases repetition or lowers ROUGE.}
\label{tab:discrete_tuning}
\resizebox{\columnwidth}{!}{%
\begin{tabular}{ll rrrrr}
\toprule
Task & Config & Temp & Len & Rep/Degen & Score & sec/ex \\
\midrule
Open & B=256, default & 0.5 & 31.7 & 9.3 Rep & 0.646 D2 & 0.64 \\
Open & B=256, default & 0.0 & 32.1 & 17.9 Rep & 0.508 D2 & 0.69 \\
Open & B=256, EOS-inf & 0.5 & 137.0 & 52.5 Rep & 0.291 D2 & 0.64 \\
Open & B=32, EOS-inf & 0.0 & 163.3 & 85.8 Rep & 0.084 D2 & 0.61 \\
\midrule
XSum & B=256, default & 0.5 & 17.2 & 0.0 Deg & 24.33 R-1 & 1.23 \\
XSum & B=64, default & 0.0 & 17.4 & 0.0 Deg & 23.84 R-1 & 1.13 \\
XSum & B=256, EOS-inf & 0.5 & 85.6 & 0.0 Deg & 14.44 R-1 & 1.24 \\
XSum & B=64, EOS-inf & 0.0 & 63.7 & 6.0 Deg & 13.71 R-1 & 1.13 \\
\midrule
CNN/DM & B=256, default & 0.5 & 20.3 & 0.0 Deg & 22.36 R-1 & 1.98 \\
CNN/DM & B=256, default & 0.0 & 20.0 & 0.0 Deg & 21.96 R-1 & 1.83 \\
CNN/DM & B=256, EOS-inf & 0.5 & 159.4 & 2.0 Deg & 16.85 R-1 & 1.98 \\
CNN/DM & B=64, default & 0.5 & 49.6 & 0.0 Deg & 16.26 R-1 & 2.01 \\
\bottomrule
\end{tabular}}
\end{table}

The sweep does not reveal a hidden discrete configuration that removes the tradeoff. The best ROUGE settings are very short full-sequence outputs, while EOS-forced settings better match target length but reintroduce repetition or degenerate endings. We therefore keep LLaDA+EOS+Block as an engineered comparison, not as a claimed optimum over all possible discrete samplers.

\subsection{BERTScore}
\label{app:sum_bertscore}

Table~\ref{tab:sum_bertscore} reports BERTScore (roberta-large, rescaled with baseline) for each (dataset, method) at NFE=64. BERTScore is monotone-consistent with ROUGE: SDE leads on XSum/CNN/DM; LLaDA leads on PubMed/arXiv. The rescaled values are negative on long-reference scientific datasets because the rescaling baseline (semantic similarity of random pairs) is high in those domains.

\begin{table}[ht!]
\centering
\caption{BERTScore F1 at NFE=64 (1{,}000 samples, roberta-large with rescale-with-baseline).}
\label{tab:sum_bertscore}
\resizebox{\columnwidth}{!}{%
\begin{tabular}{l rrr}
\toprule
Dataset & DSL-LLaDA-SDE & LLaDA & LLaDA+EOS+Block \\
\midrule
XSum    & \best{34.08} & 28.78 & 19.41 \\
CNN/DM  & \best{14.33} &  9.44 &  2.10 \\
PubMed  & $-$1.72 & \best{2.15} & $-$6.45 \\
arXiv   & $-$7.15 & \best{$-$0.62} & $-$11.76 \\
\bottomrule
\end{tabular}}
\end{table}

\subsection{Automated failure-mode tagging}
\label{app:sum_failures}

Table~\ref{tab:sum_failures} reports per-sample failure-mode rates. We tag a sample as \textbf{premature\_eos} if its word count is below 40\% of the reference; \textbf{severe phrase repetition} if the most-frequent 4-gram covers $\geq$15\% of all 4-grams (distinct from the word-level consecutive-repetition rate ``Rep'' in \S\ref{sec:sde}); \textbf{degen\_ending} if the last 40 characters contain $\geq$8 periods or commas, or the text contains ``.......''. The most striking contrast is that LLaDA+EOS+Block produces degenerate endings on 23.8\% of XSum outputs and 7.1\% of CNN/DM outputs, while DSL-LLaDA-SDE is at 0\% and 3.6\% on the same datasets. Vanilla LLaDA's failure mode is opposite: 82\% of XSum outputs and 92\% of CNN/DM outputs are \emph{below} 40\% of the reference length, reflecting the EOS cascade.

\begin{table}[ht!]
\centering
\caption{Automated failure-mode rates (\% of 1{,}000 samples, NFE=64).}
\label{tab:sum_failures}
\resizebox{\columnwidth}{!}{%
\begin{tabular}{l rrr rrr rrr}
\toprule
& \multicolumn{3}{c}{DSL-LLaDA-SDE} & \multicolumn{3}{c}{LLaDA} & \multicolumn{3}{c}{LLaDA+EOS+Block} \\
\cmidrule(lr){2-4}\cmidrule(lr){5-7}\cmidrule(lr){8-10}
Dataset & preEOS & 4g-rep & degen & preEOS & 4g-rep & degen & preEOS & 4g-rep & degen \\
\midrule
XSum   & 14.9 &  2.9 & \best{0.0} & 82.3 & 22.1 & \best{0.0} & 0.3 &  1.3 & 23.8 \\
CNN/DM & 0.9  & 14.3 & 3.6        & 91.6 & 17.9 & \best{0.0} & 1.9 & 13.8 &  7.1 \\
PubMed & 5.2  &  6.7 & 0.2        & 1.7  &  1.6 & 0.1        & 4.1 &  6.4 & 0.6  \\
arXiv  & 1.0  & 11.5 & 0.4        & 0.8  &  1.4 & 0.0        & 4.2 & 11.3 & 0.1  \\
\bottomrule
\end{tabular}}
\end{table}

\subsection{LLM-as-judge: pairwise preference and per-axis means}
\label{app:judge_axes}

Table~\ref{tab:sum_quality} reports the pairwise preference rates between SDE and the two discrete baselines at NFE=16 and NFE=64.

\begin{table}[ht!]
\centering
\caption{LLM-as-judge pairwise preference rate (\%, 100 samples/dataset, GPT-5.4) at NFE=16 and NFE=64. SDE wins in aggregate against vanilla LLaDA at NFE=16 and decisively against LLaDA$^\star$ at both budgets; vanilla LLaDA matches or beats SDE on long-reference targets at NFE=64.}
\label{tab:sum_quality}
\resizebox{\columnwidth}{!}{%
\begin{tabular}{l rr rr rr rr}
\toprule
& \multicolumn{4}{c}{\textbf{vs.\ LLaDA}} & \multicolumn{4}{c}{\textbf{vs.\ LLaDA$^\star$}} \\
\cmidrule(lr){2-5}\cmidrule(lr){6-9}
& \multicolumn{2}{c}{NFE=16} & \multicolumn{2}{c}{NFE=64} & \multicolumn{2}{c}{NFE=16} & \multicolumn{2}{c}{NFE=64} \\
\cmidrule(lr){2-3}\cmidrule(lr){4-5}\cmidrule(lr){6-7}\cmidrule(lr){8-9}
Dataset & SDE & base & SDE & base & SDE & base & SDE & base \\
\midrule
XSum    & \best{66.3} & 21.7 & \best{51.0} & 39.0        & \best{87.5} & 3.8 & \best{61.0} & 30.0 \\
CNN/DM  & 45.7        & 46.9 & \best{58.8} & 26.8        & \best{90.5} & 2.4 & \best{59.1} & 23.7 \\
PubMed  & \best{56.6} & 22.9 & 45.0        & \best{51.0} & \best{84.1} & 0.0 & \best{59.0} & 33.0 \\
arXiv   & \best{63.0} & 14.8 & 32.0        & \best{61.0} & \best{72.2} & 0.0 & \best{60.0} & 31.0 \\
\midrule
\textit{mean} & \best{57.9} & 26.6 & \best{46.7} & 44.5 & \best{83.6} & 1.5 & \best{59.8} & 29.4 \\
\bottomrule
\end{tabular}}
\end{table}

Table~\ref{tab:judge_axes} reports the per-axis means (1--5 scale) from GPT-5.4 pairwise judgment, comparing DSL-LLaDA-SDE against vanilla LLaDA. The pattern follows the regime split in the main text: on XSum and CNN/DM, SDE wins fluency and coverage; on PubMed and arXiv, LLaDA wins coverage and factuality because its longer outputs cover more reference content (without the EOS-cascade collapse seen on shorter references).

\begin{table}[ht!]
\centering
\caption{LLM-as-judge per-axis means: DSL-LLaDA-SDE vs.\ vanilla LLaDA (100 samples/dataset, GPT-5.4, 1--5 scale).}
\label{tab:judge_axes}
\resizebox{\columnwidth}{!}{%
\begin{tabular}{l rrrr rrrr}
\toprule
& \multicolumn{4}{c}{DSL-LLaDA-SDE} & \multicolumn{4}{c}{LLaDA} \\
\cmidrule(lr){2-5}\cmidrule(lr){6-9}
Dataset & fact & cov & flu & conc & fact & cov & flu & conc \\
\midrule
XSum   & 2.83 & \best{2.86} & \best{4.65} & 4.22 & \best{2.89} & 2.44 & 3.86 & \best{4.29} \\
CNN/DM & 3.09 & \best{3.22} & 3.08        & 2.85 & \best{3.14} & 1.57 & \best{3.40} & \best{4.40} \\
PubMed & 2.50 & 2.74 & \best{2.70} & 2.49 & \best{2.80} & \best{3.05} & 2.53 & \best{2.60} \\
arXiv  & 2.33 & 2.36 & 2.25 & 2.12 & \best{2.88} & \best{3.07} & \best{2.51} & \best{2.60} \\
\bottomrule
\end{tabular}}
\end{table}

\subsection{Judge reliability checks}
\label{app:judge_reliability}

Table~\ref{tab:judge_reliability} reports bootstrap 95\% CIs and position-bias checks (shown-A) for representative comparisons.

\begin{table}[ht!]
\centering
\caption{Representative GPT-5.4 judge reliability checks. Win rates report bootstrap 95\% CIs over judged samples; shown-A is the fraction of non-tie judgments selecting the answer displayed first. Large shown-A deviations are flagged as residual position-bias risk.}
\label{tab:judge_reliability}
\resizebox{\columnwidth}{!}{%
\begin{tabular}{llr rrr r}
\toprule
Dataset & Comparison & NFE & SDE win & Other win & 95\% CI (SDE) & shown-A \\
\midrule
XSum & SDE vs LLaDA & 16 & 66.3 & 21.7 & [56.6, 75.9] & 53.4 \\
CNN/DM & SDE vs LLaDA & 16 & 45.7 & 46.9 & [34.6, 56.8] & 58.7 \\
PubMed & SDE vs LLaDA & 16 & 56.6 & 22.9 & [45.8, 67.5] & 54.5 \\
arXiv & SDE vs LLaDA & 16 & 63.0 & 14.8 & [51.8, 72.8] & 57.1 \\
\midrule
XSum & SDE vs EOS+Block & 64 & 61.0 & 30.0 & [51.0, 70.0] & 62.6 \\
CNN/DM & SDE vs EOS+Block & 64 & 59.1 & 23.7 & [49.5, 68.8] & 46.8 \\
PubMed & SDE vs EOS+Block & 64 & 59.0 & 33.0 & [49.0, 68.0] & 39.1 \\
arXiv & SDE vs EOS+Block & 64 & 60.0 & 31.0 & [50.0, 69.0] & 49.5 \\
\midrule
XSum & SDE vs Qwen-AR & 64 & 37.0 & 53.0 & [28.0, 46.0] & 61.1 \\
CNN/DM & SDE vs Qwen-AR & 64 & 20.2 & 77.7 & [12.8, 28.7] & 47.8 \\
PubMed & SDE vs Qwen-AR & 64 & 31.6 & 66.3 & [22.1, 41.0] & 49.5 \\
arXiv & SDE vs Qwen-AR & 64 & 21.9 & 71.9 & [13.5, 30.2] & 56.7 \\
\bottomrule
\end{tabular}}
\end{table}

Most shown-A values stay within $\pm$10 points of 50\%; XSum shows residual first-position preference in some comparisons. We interpret judge results as supporting evidence rather than ground truth.

\subsection{Qualitative case studies}
\label{app:sum_cases}

To show what the per-sample numbers mean in practice, we present one representative comparison per dataset, sampled from the top of the SDE-wins bucket (large $+\Delta$R-1) or loses bucket. All outputs are exactly what the model produced; no editing was applied.

\paragraph{XSum~(id=195, $\Delta$R-1=+60).} Reference (15w): \textit{``Worcestershire's Daryl Mitchell has been elected as the new chairman of the Professional Cricketers' Association.''} SDE (18w): \textit{``Worcestershire captain Daryl Mitchell has been elected as the new chairman of the Professional Cricket Players Association (PCA).''} LLaDA+EOS+Block (91w): \textit{``D Worcestershire opener Daryl Mitchell has been elected as the new chairman of the Professional Cricket Players Association (PCA), taking over from Markorgan wicketkeeper Mark Wallace. [\ldots] Mitchell will be the third Worcestershire captain to fulfill the role, following Tim Curtis and Vikram Solanki.''} SDE matches reference length and content; Block fills the 128-token budget with fabricated biographical details (``Markorgan wicketkeeper'' is not in the source).

\paragraph{CNN/DM~(id=853, $\Delta$R-1=+64).} Reference (35w): \textit{``Jenrry Mejia suspended for 80 New York Mets games without pay. Mejia tested positive for banned substance stanozolol. [\ldots]''} SDE (36w): \textit{``New York Mets closer Jenrry Mejia was suspended for 80 games without pay after testing positive for the banned substance stanozolol. The 25-year-old right hander will not be able to play again until [\ldots]''} Block (189w) enters repetition: \textit{``[\ldots] Mejia, he is unaware of how the substance entered his body. Mejia, Mejia, [\ldots]''}. SDE delivers a clean 36-word news-lead; Block pads to 189 words and enters a repetitive loop on the subject's name.

\paragraph{arXiv~(id=537, $\Delta$R-1=+60).} Reference is a technical abstract on radio-frequency-dressed atom traps. SDE (210w) produces a coherent abstract-style summary: \textit{``The paper discusses the use of radio frequency dressed potentials (rf-dressed potentials) to trap and manipulate ultra-cold atoms [\ldots]''}. Block (12w) collapses into repetition of a single token after 10 words: \textit{``The paper paper potentials, proposed by Zobayay Garraway, are used to manipulate\underline{old}\underline{old}\underline{old}\underline{old}[\ldots]''}. This is the severe block-remasking failure mode on long scientific prompts.

\paragraph{arXiv~(id=491, $\Delta$R-1=$-$37).} A representative case where SDE loses on a long-reference scientific abstract. Reference is a 273-word abstract on a $\gamma$-ray quasar 4C+55.17 observed by Fermi-LAT. SDE (206w) opens coherently but enters a repetition loop on a high-energy acronym: \textit{``[\ldots] MAGIC VHE\underline{HE}\underline{HE}\underline{HE}\underline{HE}\underline{HE}\underline{HE}\ldots''}. Block (179w) maintains a coherent abstract structure: \textit{``Gamma-ray astronomy has grown rapidly due to the development of High Energy Cherenkov telescopes (IHECTS) [\ldots]''}. On long, technical-acronym-heavy text, the block-remasking schedule's commitment rhythm helps the model avoid the semantic-attractor repetition that SDE's parallel updates can fall into on extended sequences.

\subsection{Autoregressive reference baseline}
\label{app:ar_baseline}

To put the diffusion results in context, we run \textbf{Qwen2.5-7B-Instruct} (the closest publicly available 7--8B AR LLM in our local cache) on the same 1{,}000-sample test sets, with greedy decoding, the same chat-template prompts, and matched generation lengths. AR is a different paradigm and a strong reference, not a like-for-like comparison; we report it to ground the diffusion numbers and to characterize the wall-time cost of each inference mode.

\paragraph{ROUGE.} Table~\ref{tab:ar_rouge} reports R-1/R-2/R-L for all four datasets against both AR references. Mean R-1: 38.5 (Llama-3.1-8B), 37.0 (Qwen2.5-7B), 35.7 (DSL-LLaDA-SDE NFE=64), 31.5 (LLaDA NFE=64). SDE exceeds both AR references by 4--6 R-1 on XSum and is within 1--10 points elsewhere; the two AR baselines agree within 0.0--2.7 points except on PubMed (Llama-3 leads by 2.7) and arXiv (Llama-3 leads by 2.2). Vanilla LLaDA at NFE=64 sits between SDE and the AR references on long-reference benchmarks. The 8B-scale diffusion approach is therefore competitive with strong AR baselines on this surface metric, despite running parallel rather than sequential decoding.

\begin{table}[ht!]
\centering
\caption{ROUGE F1 vs.\ AR baselines and public XDLM Base (1{,}000 samples per benchmark, seed=42). Mean R-1: Llama-3.1 38.5, Qwen2.5 37.0, SDE 35.7, LLaDA 31.5, XDLM Base 18.0. XDLM Base + xDM uses the official released checkpoint with its native sampler ($k_1{=}0.1$, plain prompt); all others use chat-template prompts.}
\label{tab:ar_rouge}
\resizebox{\columnwidth}{!}{%
\begin{tabular}{ll rrr}
\toprule
Dataset & Method & R-1 & R-2 & R-L \\
\midrule
\multirow{5}{*}{XSum (21w)}
& \best{DSL-LLaDA-SDE} (NFE=64) & \best{32.4} & \best{11.0} & \best{25.0} \\
& Llama-3.1-8B AR               & 27.8        & 8.0         & 19.7 \\
& Qwen2.5-7B AR                 & 26.6        & 6.9         & 19.1 \\
& LLaDA (NFE=64)                & 24.8        & 7.7         & 20.6 \\
& XDLM Base + xDM (NFE=64)      & 15.4        & 4.3         & 11.9 \\
\midrule
\multirow{5}{*}{CNN/DM (55w)}
& Llama-3.1-8B AR               & \best{37.8} & 14.3        & 23.6 \\
& Qwen2.5-7B AR                 & \best{37.8} & 13.1        & 23.6 \\
& DSL-LLaDA-SDE (NFE=64)        & 35.8        & \best{14.9} & \best{24.3} \\
& LLaDA (NFE=64)                & 18.2        & 6.5         & 14.4 \\
& XDLM Base + xDM (NFE=64)      & 23.0        & 9.5         & 15.8 \\
\midrule
\multirow{5}{*}{PubMed (205w)}
& \best{Llama-3.1-8B AR}        & \best{45.1} & \best{16.5} & \best{24.5} \\
& Qwen2.5-7B AR                 & 42.4        & 14.2        & 22.6 \\
& LLaDA (NFE=64)                & 42.2        & 15.8        & 23.2 \\
& DSL-LLaDA-SDE (NFE=64)        & 39.5        & 14.9        & 22.8 \\
& XDLM Base + xDM (NFE=64)      & 18.9        & 4.9         & 12.4 \\
\midrule
\multirow{5}{*}{arXiv (163w)}
& \best{Llama-3.1-8B AR}        & \best{43.5} & \best{15.3} & \best{23.4} \\
& Qwen2.5-7B AR                 & 41.3        & 13.1        & 21.2 \\
& LLaDA (NFE=64)                & 40.8        & 14.3        & 21.7 \\
& DSL-LLaDA-SDE (NFE=64)        & 34.9        & 12.4        & 20.5 \\
& XDLM Base + xDM (NFE=64)      & 14.6        & 2.6         & 11.2 \\
\bottomrule
\end{tabular}}
\end{table}

\paragraph{LLM-as-judge against AR baselines.} GPT-5.4 strongly prefers both AR baselines' outputs to SDE's (Table~\ref{tab:ar_judge}): against Qwen2.5-7B, mean preference is 67\%/28\%/5\% (AR/SDE/tie) at NFE=64 and 91\%/6\%/3\% at NFE=16. Llama-3.1-8B is preferred even more decisively (73\%/22\%/5\% at NFE=64; 93\%/4.5\%/2.6\% at NFE=16), consistent with its 1.5-point ROUGE lead. The gap is driven primarily by fluency: at the same scale, AR remains the stronger generator under modern LLM evaluators, and DSL-LLaDA does not close that gap. We do not claim quality parity with AR; what diffusion provides is a different mechanism (parallel updates, error-correction-aware training) that wins decisively against the discrete-diffusion alternative (Section~\ref{sec:correction}, \S\ref{sec:robustness}) and that exhibits a complementary wall-time profile (below).

\begin{table}[ht!]
\centering
\caption{LLM-as-judge pairwise preference against AR baselines (100 samples/dataset, GPT-5.4). Both AR models are preferred over SDE on every dataset; Llama-3.1 is preferred more decisively than Qwen2.5. AR is the natural quality reference for 7--8B models; we do not claim parity.}
\label{tab:ar_judge}
\resizebox{\columnwidth}{!}{%
\begin{tabular}{l rrr rrr rrr rrr}
\toprule
& \multicolumn{6}{c}{\textbf{vs.\ Qwen2.5-7B AR}} & \multicolumn{6}{c}{\textbf{vs.\ Llama-3.1-8B AR}} \\
\cmidrule(lr){2-7}\cmidrule(lr){8-13}
& \multicolumn{3}{c}{NFE=16} & \multicolumn{3}{c}{NFE=64} & \multicolumn{3}{c}{NFE=16} & \multicolumn{3}{c}{NFE=64} \\
\cmidrule(lr){2-4}\cmidrule(lr){5-7}\cmidrule(lr){8-10}\cmidrule(lr){11-13}
Dataset & SDE & AR & Tie & SDE & AR & Tie & SDE & AR & Tie & SDE & AR & Tie \\
\midrule
XSum    & 21.0 & 73.0 & 6.0  & 37.0 & 53.0 & 10.0 & 13.2 & 80.3 & 6.6 & 32.0 & 62.7 & 5.3 \\
CNN/DM  &  4.1 & 94.8 & 1.0  & 20.2 & 77.7 &  2.1 &  3.6 & 94.0 & 2.4 &  9.7 & 86.1 & 4.2 \\
PubMed  &  0.0 & 96.9 & 3.1  & 31.6 & 66.3 &  2.1 &  1.3 & 97.3 & 1.3 & 30.7 & 65.3 & 4.0 \\
arXiv   &  0.0 & 98.9 & 1.1  & 21.9 & 71.9 &  6.2 &  0.0 &100.0 & 0.0 & 15.6 & 79.2 & 5.2 \\
\midrule
\textit{mean} & 6.3 & 90.9 & 2.8 & 27.7 & 67.2 & 5.1 & 4.5 & 92.9 & 2.6 & 22.0 & 73.3 & 4.7 \\
\bottomrule
\end{tabular}}
\end{table}

\paragraph{Wall-time benchmark.} We measure end-to-end inference time per sample on a single H100 (80GB), 20 prompts per dataset (80 total), single GPU, model loaded once. Qwen-AR is run with KV cache enabled (default) and disabled (to characterize the structural cost of incremental decoding without caching); SDE is run at NFE $\in \{8, 16, 32, 64\}$. Wall times are reported in Table~\ref{tab:wall_time}.

\begin{table}[ht!]
\centering
\caption{Wall time per sample (median seconds, single H100), 80 prompts (20/dataset). Qwen-AR and Llama-3.1-AR use KV cache by default; with cache disabled they scale quadratically in output length. SDE uses NFE forward passes over the full sequence; cost is roughly linear in NFE. The two AR baselines have similar wall-time profiles; Llama-3.1 is slightly slower due to longer average outputs.}
\label{tab:wall_time}
\resizebox{\columnwidth}{!}{%
\begin{tabular}{l rrrr r}
\toprule
Method & XSum & CNN/DM & PubMed & arXiv & \textit{mean} \\
\midrule
SDE NFE=8                & 0.76 & 1.32 &  2.21 &  2.24 & \textit{1.63} \\
SDE NFE=16               & 1.38 & 2.38 &  3.97 &  4.02 & \textit{2.94} \\
Qwen-AR (KV on)          & 1.36 & 2.64 &  7.13 &  7.25 & \textit{4.59} \\
Llama-3.1-AR (KV on)     & 1.80 & 3.89 &  7.95 &  6.75 & \textit{5.10} \\
SDE NFE=32               & 2.44 & 4.22 &  7.07 &  7.14 & \textit{5.22} \\
SDE NFE=64               & 4.87 & 8.44 & 14.14 & 14.27 & \textit{10.43} \\
Qwen-AR (KV off)         & 2.54 & 7.19 & 36.97 & 40.30 & \textit{21.75} \\
Llama-3.1-AR (KV off)    & 3.33 & 10.11 & 42.13 & 37.34 & \textit{23.23} \\
\bottomrule
\end{tabular}}
\end{table}

Two efficiency observations follow: (1)~SDE at the few-step regime is not slower than either AR baseline, as NFE=16 matches Qwen-AR / Llama-3.1-AR with KV cache on XSum and CNN/DM and is 1.7--2.0$\times$ faster on PubMed and arXiv (3.97--4.02s for SDE vs.\ 7.13--7.95s for the AR baselines), because AR's incremental decoding still carries an attention cost over the full prompt at every output token, and (2)~NFE=8 is uniformly 1.8--3.6$\times$ faster than the AR baselines at the cost of the quality drop reported above. The KV-off rows show the quadratic blow-up that AR avoids only because of the cache; SDE's continuous-noise dynamics do not admit an analogous incremental shortcut, but in exchange they fix the number of forward passes at a small NFE, which is cheaper than $L$ token-level decodes for long outputs over long prompts.

\subsection{Public XDLM Base + xDM Reference}
\label{app:xdlm_public}

For completeness, we evaluate the publicly released XDLM Base checkpoint\footnote{\url{https://huggingface.co/GSAI-ML/LLaDA-8B-XDLM}} with its native xDM sampler (\texttt{generate\_xdm}, $k_1{=}0.1$) on the same four summarization benchmarks (1{,}000 samples, seed 42, temperature 0.0, plain prompt format). This checkpoint is continual pretraining from LLaDA-8B-\emph{Base} (not the instruction-tuned variant) and the official repository evaluates it on reasoning/coding tasks (GSM8K, MATH, HumanEval, MBPP, BBH) rather than zero-shot summarization.

Table~\ref{tab:xdlm_public_r1} compares ROUGE-1 F1 across NFE budgets, placing the public XDLM Base + xDM alongside DSL-LLaDA-SDE and LLaDA from Table~\ref{tab:summarization_hero}. The public checkpoint is substantially weaker at every operating point: its best result (23.01 R-1 on CNN/DM at NFE=64) is still below LLaDA's worst (24.8 on XSum) and roughly half of DSL-LLaDA-SDE at the same budget. Qualitative inspection shows that many outputs begin with a plausible summary sentence but then drift into copied source text, dialogue-style continuations, or repetition---consistent with a base LM receiving a task prompt without instruction tuning. XDLM Base + xDM does improve substantially with more NFE (e.g., CNN/DM rises from 3.27 to 23.01), suggesting the xDM sampler functions correctly but the base checkpoint lacks the instruction-following ability needed for competitive summarization.

\paragraph{Setting differences from our re-implemented XDLM.} The XDLM baseline used throughout the main text is a matched-compute re-implementation: we continue-pretrain LLaDA-8B-Instruct with the XDLM/XDM forward process on the same data, steps, and learning rate as DSL-LLaDA (Section~\ref{sec:experiments}), then decode with discrete remasking. The public checkpoint differs in three ways: (1)~it starts from LLaDA-8B-Base rather than the instruction-tuned variant, (2)~it uses the official xDM continuous sampler rather than discrete remasking, and (3)~its training budget and data may differ from ours. We include these numbers as a reference for the published XDLM method's zero-shot summarization performance; the low scores should not be read as an indictment of the xDM sampler but as a reflection of the base-only checkpoint's limited instruction-following ability.

\begin{table}[ht!]
\centering
\caption{ROUGE-1 F1 comparison: public XDLM Base + xDM vs.\ instruction-tuned baselines from Table~\ref{tab:summarization_hero} (1{,}000 samples, seed=42). The public checkpoint uses the native xDM sampler ($k_1{=}0.1$, plain prompt); DSL-LLaDA-SDE and LLaDA use the same chat-template prompts and settings as the main text. The large gap reflects the lack of instruction tuning in the public XDLM checkpoint, not a sampler limitation.}
\label{tab:xdlm_public_r1}
\footnotesize
\setlength{\tabcolsep}{6pt}
\renewcommand{\arraystretch}{1.05}
\resizebox{\columnwidth}{!}{%
\begin{tabular}{ll cccc}
\toprule
Dataset (ref) & Method & NFE=8 & 16 & 32 & 64 \\
\midrule
\multirow{3}{*}{XSum (21w)}
& DSL-LLaDA-SDE          & \best{28.4} & \best{30.4} & \best{32.0} & \best{32.4} \\
& LLaDA                   & 25.2        & 29.0        & 28.8        & 24.8        \\
& XDLM Base + xDM         &  4.5        &  7.6        & 14.4        & 15.4        \\
\midrule
\multirow{3}{*}{CNN/DM (55w)}
& DSL-LLaDA-SDE          & \best{28.1} & \best{33.0} & \best{35.1} & \best{35.8} \\
& LLaDA                   & 23.1        & 28.2        & 25.4        & 18.2        \\
& XDLM Base + xDM         &  3.3        &  4.6        & 11.0        & 23.0        \\
\midrule
\multirow{3}{*}{PubMed (205w)}
& DSL-LLaDA-SDE          & \best{29.4} & \best{32.2} & \best{36.9} & 39.5        \\
& LLaDA                   & 11.9        & 20.2        & 36.5        & \best{42.2} \\
& XDLM Base + xDM         &  7.5        &  7.7        & 10.9        & 18.9        \\
\midrule
\multirow{3}{*}{arXiv (163w)}
& DSL-LLaDA-SDE          & \best{27.3} & \best{28.6} & 32.9        & 34.9        \\
& LLaDA                   & 10.6        & 16.2        & \best{33.8} & \best{40.8} \\
& XDLM Base + xDM         &  6.0        &  6.3        &  8.8        & 14.6        \\
\bottomrule
\end{tabular}}
\end{table}

Table~\ref{tab:xdlm_public_full} provides the full ROUGE-2, ROUGE-L, and average output length for the public XDLM Base + xDM checkpoint.

\begin{table}[ht!]
\centering
\caption{Full metrics for public XDLM Base + xDM (1{,}000 samples, seed=42, plain prompt, $k_1{=}0.1$).}
\label{tab:xdlm_public_full}
\resizebox{\columnwidth}{!}{%
\begin{tabular}{l r rrr r}
\toprule
Dataset & NFE & ROUGE-1 & ROUGE-2 & ROUGE-L & Avg.\ words \\
\midrule
\multirow{4}{*}{XSum}
& 8  &  4.52 & 0.60 &  4.02 & 103.30 \\
& 16 &  7.55 & 1.57 &  6.41 &  94.32 \\
& 32 & 14.37 & 4.01 & 11.36 &  87.09 \\
& 64 & 15.41 & 4.34 & 11.88 &  90.92 \\
\midrule
\multirow{4}{*}{CNN/DM}
& 8  &  3.27 & 0.40 &  2.83 & 210.23 \\
& 16 &  4.62 & 0.86 &  3.86 & 196.35 \\
& 32 & 10.97 & 3.74 &  7.87 & 182.19 \\
& 64 & 23.01 & 9.54 & 15.82 & 183.79 \\
\midrule
\multirow{4}{*}{PubMed}
& 8  &  7.45 & 0.94 &  5.44 & 190.65 \\
& 16 &  7.71 & 1.12 &  5.57 & 192.24 \\
& 32 & 10.88 & 2.31 &  7.48 & 200.78 \\
& 64 & 18.85 & 4.87 & 12.42 & 198.87 \\
\midrule
\multirow{4}{*}{arXiv}
& 8  &  5.95 & 0.26 &  5.45 & 204.75 \\
& 16 &  6.27 & 0.46 &  5.65 & 204.46 \\
& 32 &  8.83 & 1.21 &  7.36 & 197.17 \\
& 64 & 14.55 & 2.55 & 11.21 & 187.14 \\
\bottomrule
\end{tabular}}
\end{table}

\section{Full Generation Quality Comparison}
\label{app:gen_quality}

Table~\ref{tab:gen_full} reports all methods and NFE budgets tested (200 prompts, gen=256). Three EOS suppression variants are included: LLaDA+EOS (EOS logit forced to $-\infty$), LLaDA+EOS+Block (block remasking with EOS suppression), and DSL-LLaDA-Remask+EOS (DSL model with EOS suppression). All three degenerate into high repetition, confirming that SDE's advantage comes from continuous dynamics, not from suppressing EOS.

\begin{table}[ht!]
\centering
\caption{Complete generation quality comparison (200 prompts, gen=256). $\dagger$: GenPPL unreliable at Rep${>}$30\%.}
\label{tab:gen_full}
\resizebox{\columnwidth}{!}{%
\begin{tabular}{rl cccr}
\toprule
NFE & Method & GenPPL$\downarrow$ & D2$\uparrow$ & Rep$\downarrow$ & Len$\uparrow$ \\
\midrule
\multirow{7}{*}{8}
& LLaDA (default) & 63.1 & \best{0.470} & 0.248 & 27 \\
& LLaDA+EOS & 7.2$^\dagger$ & 0.132 & 0.777 & 113 \\
& LLaDA+EOS+Block & 6.3$^\dagger$ & 0.064 & 0.895 & 116 \\
& DSL-LLaDA-Remask & 12.5$^\dagger$ & 0.164 & 0.663 & 92 \\
& DSL-LLaDA-Remask+EOS & 3.2$^\dagger$ & 0.064 & 0.855 & 183 \\
& DSL-LLaDA-Remask+Block & 3.6$^\dagger$ & 0.036 & 0.930 & 171 \\
& \textbf{DSL-LLaDA-SDE} & \best{12.9} & 0.342 & \best{0.083} & \best{174} \\
\midrule
\multirow{7}{*}{16}
& LLaDA (default) & 39.1 & \best{0.474} & 0.216 & 32 \\
& LLaDA+EOS & 7.5$^\dagger$ & 0.225 & 0.593 & 123 \\
& LLaDA+EOS+Block & 5.1$^\dagger$ & 0.086 & 0.845 & 143 \\
& DSL-LLaDA-Remask & 11.4$^\dagger$ & 0.221 & 0.600 & 92 \\
& DSL-LLaDA-Remask+EOS & 3.7$^\dagger$ & 0.117 & 0.656 & 184 \\
& DSL-LLaDA-Remask+Block & 2.9$^\dagger$ & 0.046 & 0.918 & 211 \\
& \textbf{DSL-LLaDA-SDE} & \best{8.3} & 0.299 & \best{0.087} & \best{174} \\
\midrule
\multirow{6}{*}{32}
& LLaDA (default) & 20.3 & \best{0.504} & 0.168 & 45 \\
& LLaDA+EOS & 7.0$^\dagger$ & 0.310 & 0.353 & 156 \\
& LLaDA+EOS+Block & 5.3$^\dagger$ & 0.140 & 0.718 & 163 \\
& DSL-LLaDA-Remask & 10.7$^\dagger$ & 0.291 & 0.304 & 119 \\
& DSL-LLaDA-Remask+Block & 3.5$^\dagger$ & 0.095 & 0.774 & 206 \\
& \textbf{DSL-LLaDA-SDE} & \best{6.1} & 0.322 & \best{0.064} & \best{185} \\
\midrule
\multirow{8}{*}{64}
& LLaDA (default) & 12.8 & \best{0.489} & 0.095 & 62 \\
& LLaDA+EOS & 8.1 & 0.422 & 0.053 & 194 \\
& LLaDA+EOS+Block & 7.6 & 0.305 & 0.200 & 182 \\
& LLaDA+Block & 7.6 & 0.305 & 0.227 & 188 \\
& DSL-LLaDA-Remask & 9.6 & 0.405 & 0.058 & 140 \\
& DSL-LLaDA-Remask+EOS & 6.8 & 0.379 & 0.104 & 193 \\
& DSL-LLaDA-Remask+Block & 6.8 & 0.243 & 0.244 & 200 \\
& \textbf{DSL-LLaDA-SDE} & \best{5.1} & 0.334 & \best{0.037} & \best{187} \\
\midrule
\multirow{8}{*}{128}
& LLaDA (default) & 11.5 & \best{0.512} & \best{0.003} & 49 \\
& LLaDA+EOS & 6.7 & 0.459 & 0.009 & 198 \\
& LLaDA+EOS+Block & 6.8 & 0.408 & 0.015 & 179 \\
& LLaDA+Block & 6.8 & 0.397 & 0.014 & 204 \\
& DSL-LLaDA-Remask & 7.5 & 0.421 & 0.009 & 142 \\
& DSL-LLaDA-Remask+EOS & 6.2 & 0.423 & 0.019 & 201 \\
& DSL-LLaDA-Remask+Block & 7.0 & 0.361 & 0.017 & \best{208} \\
& \textbf{DSL-LLaDA-SDE} & \best{4.5} & 0.300 & 0.075 & 154 \\
\bottomrule
\end{tabular}}
\end{table}

Key observations: (1)~All EOS suppression variants (LLaDA+EOS, LLaDA+EOS+Block) trade short output for massive repetition: LLaDA+EOS+Block achieves GenPPL$=$5.1 at NFE$=$16 but 85\% repetition. (2)~The $\beta{=}1$ model with eos\_inf degenerates just as badly (66--86\% repetition at NFE$\leq$16), confirming that the DSL-trained model's iterative-unmasking failure is not an EOS artifact. (3)~SDE is the only method that achieves low PPL, low repetition, and long output simultaneously at every NFE budget.

\subsection{Long-form generation}
\label{app:longform}

Table~\ref{tab:longform} reports all five methods at gen=512 and gen=1024 (100 prompts, NFE=64). Iterative unmasking's repetition scales with generation length; SDE degrades gracefully. LLaDA ReMDM produces very short outputs (50--69 words) with high repetition, indicating that confidence-based remasking does not alleviate the length--quality tradeoff at long generation horizons.

\begin{table}[ht!]
\centering
\caption{Long-form generation (100 prompts, NFE=64). D2 = fraction of unique bigrams (higher = more diverse). $\dagger$: GenPPL unreliable (Rep${>}$30\%).}
\label{tab:longform}
\resizebox{\columnwidth}{!}{%
\begin{tabular}{rl cccc}
\toprule
gen & Method & GenPPL$\downarrow$ & D2$\uparrow$ & Rep$\downarrow$ & Len$\uparrow$ \\
\midrule
\multirow{5}{*}{512}
& LLaDA+EOS & 4.6$^\dagger$ & 0.107 & 0.686 & 330 \\
& LLaDA+EOS+Block & 3.5$^\dagger$ & 0.056 & 0.891 & 371 \\
& LLaDA ReMDM & 19.3$^\dagger$ & 0.414 & 0.351 & 50 \\
& DSL-LLaDA-Remask & 8.6$^\dagger$ & 0.159 & 0.434 & 206 \\
& \textbf{DSL-LLaDA-SDE} & 5.4 & \best{0.212} & \best{0.111} & 292 \\
\midrule
\multirow{5}{*}{1024}
& LLaDA+EOS & 4.2$^\dagger$ & 0.034 & 0.827 & 608 \\
& LLaDA+EOS+Block & 3.3$^\dagger$ & 0.022 & 0.965 & 663 \\
& LLaDA ReMDM & 31.9$^\dagger$ & 0.230 & 0.647 & 69 \\
& DSL-LLaDA-Remask & 6.3$^\dagger$ & 0.034 & 0.649 & 580 \\
& \textbf{DSL-LLaDA-SDE} & 5.6 & \best{0.139} & \best{0.284} & 451 \\
\bottomrule
\end{tabular}}
\end{table}

\subsection{Generation Diversity}
\label{app:diversity}

Iterative unmasking with temperature$=$0 is fully deterministic; SDE, by contrast, injects stochastic noise at every step and produces different outputs on each run. On 20 prompts at NFE$=$64 across 5 random seeds, LLaDA and DSL-LLaDA-Remask yield bit-identical outputs (PPL constant at 9.01 and 9.63 respectively to two decimal places), while DSL-LLaDA produces PPL in the range 5.98--6.99, reflecting genuinely different generated text. This stochastic diversity is a direct consequence of continuous dynamics and is unavailable to iterative unmasking at zero temperature.

\subsection{Block SDE: Untapped Inference Headroom}
\label{app:block_sde}

The main paper uses pure SDE, which updates all $L$ positions in parallel and is best suited to tasks with strong informational anchors. We separately ask whether the same DSL-trained backbone has additional inference headroom on tasks where sequential dependency dominates and parallel updates struggle. The natural inference variant is \emph{Block SDE}: run SDE within $B$-token blocks left-to-right, committing each block before proceeding to the next. This re-introduces left-to-right credit assignment while keeping the within-block continuous denoising that DSL training was designed for.

Table~\ref{tab:block_sweep} reports the GSM8K development sweep used to select the Block-SDE schedule (100 problems, gen=256, seed=42, regex answer extraction). The sweep includes three reference points: full-sequence discrete remasking ($B{=}256$, 40\%), discrete block remasking ($B{=}32$, 58\%), and pure SDE without blocks (28\%). Table~\ref{tab:block_full} then reports the full-set confirmation on GSM8K and MATH-500.

\begin{table}[ht!]
\centering
\caption{Development sweep for Block SDE on 100 GSM8K problems (gen=256). NFE $= (256/B) \times 2S$ (Heun solver). The 77\% result selected $B{=}8$, $S{=}4$ for the full GSM8K run, but the small development set overestimates absolute accuracy; Table~\ref{tab:block_full} gives the full-set result.}
\label{tab:block_sweep}
\renewcommand{\arraystretch}{1.1}
\resizebox{\columnwidth}{!}{%
\begin{tabular}{rr rrr}
\toprule
$B$ & $S$ & NFE & Acc. & Notes \\
\midrule
\multicolumn{2}{l}{\textit{Reference points}} \\
256 & --- & 64 & 40\% & Discrete remask (full sequence) \\
32 & --- & 64 & 58\% & Discrete block remask \\
--- & 32 & 64 & 28\% & Pure SDE (no blocks) \\
\midrule
\multicolumn{2}{l}{\textit{Block SDE}} \\
8 & 2 & 160 & 58\% & \\
\best{8} & \best{4} & \best{288} & \best{77\%} & Selected for full GSM8K \\
8 & 8 & 544 & 70\% & More steps $\not\Rightarrow$ better \\
16 & 2 & 80 & 30\% & \\
16 & 4 & 144 & 59\% & \\
16 & 8 & 272 & 72\% & \\
16 & 16 & 528 & 69\% & \\
32 & 4 & 72 & 42\% & \\
32 & 8 & 136 & 58\% & \\
32 & 16 & 264 & 56--66\% & High variance \\
64 & 16 & 132 & 57\% & \\
\bottomrule
\end{tabular}}
\end{table}

\begin{table}[ht!]
\centering
\caption{Full reasoning-set confirmation. GSM8K uses the development-selected $B{=}8$, $S{=}4$ schedule with gen=256. MATH-500 uses $B{=}32$, $S{=}16$ with gen=512. All rows use the same 1k-step DSL-LLaDA checkpoint; results are exact-match after regex/boxed-answer extraction.}
\label{tab:block_full}
\resizebox{\columnwidth}{!}{%
\begin{tabular}{llrrrr}
\toprule
Dataset & Method & NFE & Correct & Acc. (\%) & Time (s) \\
\midrule
GSM8K & Discrete block remask ($B{=}32$) & 64 & 759/1319 & 57.54 & 3512.9 \\
GSM8K & Pure SDE ($S{=}32$)              & 64 & 509/1319 & 38.59 & 5202.2 \\
GSM8K & \textbf{Block SDE} ($B{=}8$, $S{=}4$) & 288 & \best{838/1319} & \best{63.53} & 15965.4 \\
\midrule
MATH-500 & Discrete block remask ($B{=}32$) & 64 & 45/500 & 9.00 & 2067.2 \\
MATH-500 & Pure SDE ($S{=}32$)              & 64 & 83/500 & 16.60 & 3342.0 \\
MATH-500 & \textbf{Block SDE} ($B{=}32$, $S{=}16$) & 528 & \best{139/500} & \best{27.80} & 17602.1 \\
\bottomrule
\end{tabular}}
\end{table}

The full GSM8K run reduces the apparent gain from the 100-problem development sweep, but preserves the direction: Block SDE improves by +5.99 points over discrete block remasking and +24.94 over pure SDE. MATH-500 shows a larger separation, with Block SDE improving by +18.80 over remasking and +11.20 over pure SDE. The result is therefore not a new low-budget sampler claim. The improvement is evidence that the DSL-trained backbone can support a broader inference family when sequential structure is restored.

\paragraph{Implication.} The DSL-trained backbone contains reasoning capacity that pure parallel SDE cannot extract because parallel updates lack sequential structure. Restoring block-wise left-to-right scheduling improves full-set GSM8K and MATH-500, but at substantially higher NFE than the primary summarization runs. Block SDE occupies one point in a broader scheduling space: AR commits one token at a time, multi-token prediction commits several tokens per step, pure SDE updates all positions in parallel, and Block SDE commits local chunks after continuous denoising. Designing task- and model-aware schedules that choose between these regimes, or interpolate among them, is a natural next step.

\subsection{Qualitative Examples}
\label{app:examples}

We first show two paired examples at NFE$=$16 (gen$=$256), the low-step regime where Figure~\ref{fig:nfe_curve} shows the strongest contrast. We then present NFE$=$64 examples that illustrate the failure modes quantified in Table~\ref{tab:gen_full}.

\begin{figure*}[t!]
\begin{tcolorbox}[colback=gray!5, colframe=gray!50, title={Low-step qualitative comparison (NFE$=$16, gen$=$256)}, fonttitle=\bfseries\small]

\textbf{Prompt 1:} ``Write a short paragraph explaining the internet.''
\smallskip

\noindent
\begin{minipage}[t]{0.48\textwidth}
\begin{tcolorbox}[colback=red!3, colframe=red!40, title={\small LLaDA (32 words, premature termination + collapse)}, fonttitle=\small, boxsep=2pt, left=3pt, right=3pt]
\small\textit{The internet is a global network of interconnected computers that allows people to communicate and access information, access, and share data. It is a powerful tool for communication, entertainment, and, and and.}
\end{tcolorbox}
\end{minipage}\hfill
\begin{minipage}[t]{0.48\textwidth}
\begin{tcolorbox}[colback=green!3, colframe=green!40, title={\small DSL-LLaDA-SDE (200 words, full-length)}, fonttitle=\small, boxsep=2pt, left=3pt, right=3pt]
\small\textit{The internet is a global network of computers that allows people to communicate and share information, resources, and services online. It is also a network of that connects between a computer and a server that, as a web server and each other. The internet was created as a computer network, or the, in the 1960s, which was created as a way to the the internet. The server was connected to the internet, the internet as a web server, which was known as a server. [\ldots]}
\end{tcolorbox}
\end{minipage}

\medskip
\textbf{Prompt 2:} ``Explain the concept of blockchain technology.''
\smallskip

\noindent
\begin{minipage}[t]{0.48\textwidth}
\begin{tcolorbox}[colback=red!3, colframe=red!40, title={\small LLaDA (32 words, damaged fragment)}, fonttitle=\small, boxsep=2pt, left=3pt, right=3pt]
\small\textit{Blockchain technology is a decentralized, distributed ledger system that records transactions to verify transactions across a network of computers.. It is a secure of secure and that and and and and and.}
\end{tcolorbox}
\end{minipage}\hfill
\begin{minipage}[t]{0.48\textwidth}
\begin{tcolorbox}[colback=green!3, colframe=green!40, title={\small DSL-LLaDA-SDE (215 words, on-topic)}, fonttitle=\small, boxsep=2pt, left=3pt, right=3pt]
\small\textit{Blockchain technology is a distributed ledger system that allows transactions to be recorded across a network in a secure manner. It is essentially a ledger stored in a network of computers, in a form of computers, and in the, in a blockchain. Each transaction is recorded as a copy of the network. When a transaction is recorded in the form of a block. The block contains information as well as the time of the transaction and the identity of the parties [\ldots]}
\end{tcolorbox}
\end{minipage}

\medskip
\textbf{Caveat case:} ``Describe plate tectonics in simple terms.''
\smallskip

\noindent
\begin{minipage}[t]{0.48\textwidth}
\begin{tcolorbox}[colback=red!3, colframe=red!40, title={\small LLaDA (20 words, immediate collapse)}, fonttitle=\small, boxsep=2pt, left=3pt, right=3pt]
\small\textit{Plate tectonics is the process that explains Earth Earth Earth Earth Earth Earth Earth Earth.. It is the the Earth..}
\end{tcolorbox}
\end{minipage}\hfill
\begin{minipage}[t]{0.48\textwidth}
\begin{tcolorbox}[colback=yellow!5, colframe=yellow!50, title={\small DSL-LLaDA-SDE (206 words, phrase-level repetition)}, fonttitle=\small, boxsep=2pt, left=3pt, right=3pt]
\small\textit{Plate tectonics is the study of the Earth Earth's crust. It is made up of rigid plates that is broken into smaller plates called tectonic plates. These plates move on a liquid layer of the Earth called the plate the plates. The plate plate the plate plate plates plate plate plate plate plate plate plate, plate the plate is plate. [\ldots]}
\end{tcolorbox}
\end{minipage}

\smallskip
\noindent\textit{Summary:} At NFE$=$16, vanilla LLaDA exhibits premature termination (20--42 words) and word-level collapse on all three prompts. DSL-LLaDA-SDE eliminates these catastrophic failures and produces full-length answers. The plate-tectonics caveat (yellow) shows that phrase-level repetition can still occur at very low NFE.
\end{tcolorbox}
\end{figure*}

\paragraph{Higher-step examples (NFE$=$64).}

\paragraph{Prompt: ``Provide a brief overview of climate change.''}
DSL-LLaDA-SDE (191 words) produces a two-paragraph answer covering causes (fossil fuels, CO$_2$) and impacts (warming, ice melt), ending with a natural sentence. LLaDA (17 words) emits one incomplete sentence before EOS. MDM-CPT (118 words) starts coherently then degenerates into comma-and repetition (``and, and, and, and...'').

\paragraph{Prompt: ``Explain the French Revolution.''}
DSL-LLaDA-SDE (122 words) covers the Enlightenment influence, monarchy overthrow, Declaration of Rights, Republic establishment, Reign of Terror, and Napoleon's rise. XDLM (84 words) produces a decent opening paragraph followed by 28 consecutive periods. MDM-CPT (72 words) repeats ``of of the Enlightenment'' three times.

\paragraph{Prompt: ``Explain the Cold War.''}
DSL-LLaDA-SDE (119 words) covers the timeframe (1940s--1990s), ideological divide, and specific events (Korean War, Berlin Wall, nuclear proliferation). LLaDA (62 words) provides accurate but minimal information. XDLM (58 words) produces a good first half followed by 48 periods.

\paragraph{Prompt: ``Describe the Industrial Revolution.''}
DSL-LLaDA-SDE (173 words) produces a structured overview covering industrialization, manufacturing, and transportation. LLaDA (26 words) emits two sentences. MDM-CPT (102 words) begins coherently then fills the second half with ``factories factories factories factories...''.

In these examples, SDE avoids the catastrophic EOS and punctuation-spam failures seen in the discrete baselines, while every discrete baseline fails on at least one prompt. MDM-CPT achieves the lowest GenPPL (4.7) but produces the most degenerate text (44\% failure rate), illustrating why GenPPL alone is an unreliable quality metric.

SDE performs best on knowledge-grounded tasks (expository, summarization, instruction) where informational anchors prevent the parallel-update attractor problem. On creative writing, the lack of a target causes all positions to converge on high-frequency tokens. This task-dependent behavior ($\leq$4\% degenerate endings on summarization vs.\ 24\% on creative) suggests SDE is best suited for knowledge-grounded generation. The creative failure mode may be addressable via temperature sampling or hybrid SDE$\to$unmasking strategies, which we leave to future work.

Quantitative summarization results are reported in Section~\ref{sec:sde} and Table~\ref{tab:summarization_hero}.


\end{document}